\documentclass[runningheads]{llncs}

\usepackage{eccv}

\usepackage{eccvabbrv}

\usepackage{graphicx}
\usepackage{booktabs}
\usepackage{wrapfig}

\usepackage[accsupp]{axessibility}  % Improves PDF readability for those with disabilities.

\usepackage[pagebackref,breaklinks,colorlinks,citecolor=eccvblue]{hyperref}
\usepackage{orcidlink}
\usepackage{amssymb}
\usepackage{amsmath}
\usepackage{mathtools}
\usepackage{microtype}
\usepackage{graphicx}
\usepackage[normalem]{ulem}
\usepackage{subcaption}
\usepackage{booktabs} % for professional tables
\usepackage{colortbl} % for \rowcolor in tables
\usepackage[dvipsnames,svgnames,table]{xcolor}
\usepackage{fancyvrb}

\newcommand{\benchname}{MedQ-Deg\xspace}

\begin{document}

\title{\benchname: A Multidimensional Benchmark for Evaluating MLLMs Across Medical Image Quality Degradations} 

\titlerunning{\benchname: Evaluating MLLMs Under Medical Image Quality Degradations}

% % TODO FINAL: Replace with your author list. 
% % Include the authors' OCRID for the camera-ready version, if at all possible.
% Second Author\inst{2,3}\orcidlink{1111-2222-3333-4444} \and
% Third Author\inst{3}\orcidlink{2222--3333-4444-5555}}

% % TODO FINAL: Replace with an abbreviated list of authors.

% % TODO FINAL: Replace with your institution list.
% Springer Heidelberg, Tiergartenstr.~17, 69121 Heidelberg, Germany
% \url{https://github.com/liujiyaoFDU/MedQ-DEG.git} \and

\author{
Jiyao Liu\inst{1}$^{\ast}$\and
Junzhi Ning\inst{2}$^{\ast}$\and
Chenglong Ma\inst{1}$^{\ast}$\and
Wanying Qu\inst{1}\and
Jianghan Shen\inst{2}\and
Siqi Luo\inst{3}\and
Jinjie Wei\inst{1}\and 
Jin Ye\inst{2}\and
Pengze Li\inst{2}\and
Tianbin Li\inst{2}\and
Jiashi Lin\inst{2}\and\\
Hongming Shan\inst{1}\and
Xinzhe Luo\inst{4}\and
Xiaohong Liu\inst{3}\and
Lihao Liu\inst{2}\and\\ 
Junjun He\inst{2}$^{\dagger}$\and
Ningsheng Xu\inst{1}$^{\dagger}$
}

\authorrunning{J. Liu et al.}

\institute{
Fudan University, Shanghai, China \\
\email{jiyaoliu.fudan@gmail.com}
\and
Shanghai AI Lab, Shanghai, China
\and
Shanghai Jiaotong University, Shanghai, China
\and
Imperial College London, London, UK \\
Project: \href{https://uni-medical.github.io/MedQ-Robust-web}{{https://uni-medical.github.io/MedQ-Robust-web}}
}

\maketitle

\renewcommand\thefootnote{}
\footnotetext{$^\ast$Equal contribution. $^\dagger$Corresponding author. }
\renewcommand\thefootnote{\arabic{footnote}}

\begin{abstract}
  Despite impressive performance on standard benchmarks, multimodal large language models (MLLMs) face critical challenges in real-world clinical environments where medical images inevitably suffer various quality degradations. Existing benchmarks exhibit two key limitations: (1) absence of large-scale, multidimensional assessment across medical image quality gradients and (2) no systematic confidence calibration analysis.
  To address these gaps, we present \textbf{\benchname}, a comprehensive benchmark for evaluating medical MLLMs under image quality degradations. \benchname provides multi-dimensional evaluation spanning 18 distinct degradation types, 30 fine-grained capability dimensions, and 7 imaging modalities, with 24,894 question-answer pairs. Each degradation is implemented at 3 severity degrees, calibrated by expert radiologists. We further introduce \textit{Calibration Shift} metric, which quantifies the gap between a model's perceived confidence and actual performance to assess metacognitive reliability under degradation.
  Our comprehensive evaluation of 40 mainstream MLLMs reveals several critical findings: (1)~overall model performance degrades systematically as degradation severity increases, (2)~models universally exhibit the \textit{AI Dunning-Kruger Effect}, maintaining inappropriately high confidence despite severe accuracy collapse, and (3)~models display markedly differentiated behavioral patterns across capability dimensions, imaging modalities, and degradation types. 
  \textit{We hope \benchname drives progress toward medical MLLMs that are robust and trustworthy in real clinical practice.}
  
\end{abstract}

\section{Introduction}
\label{sec:intro}

    Multimodal Large Language Models (MLLMs) ~\cite{qwen3vl,singh2025gpt5,jiang2025hulu,sellergren2025medgemma}have recently demonstrated remarkable performance on medical vision-language benchmarks, in some cases approaching or even surpassing human experts~\cite{gmai2024,omnimedvqa2023}. However, these impressive results largely rely on carefully curated high-quality medical images. In real clinical environments, medical images are frequently degraded due to noise, motion artifacts, or hardware limitations, raising a critical question: can MLLMs remain reliable under such imperfect conditions?

  \begin{wrapfigure}{r}{0.48\linewidth}
    % \vspace{-2.5em}
    \centering
    \includegraphics[width=0.95\linewidth]{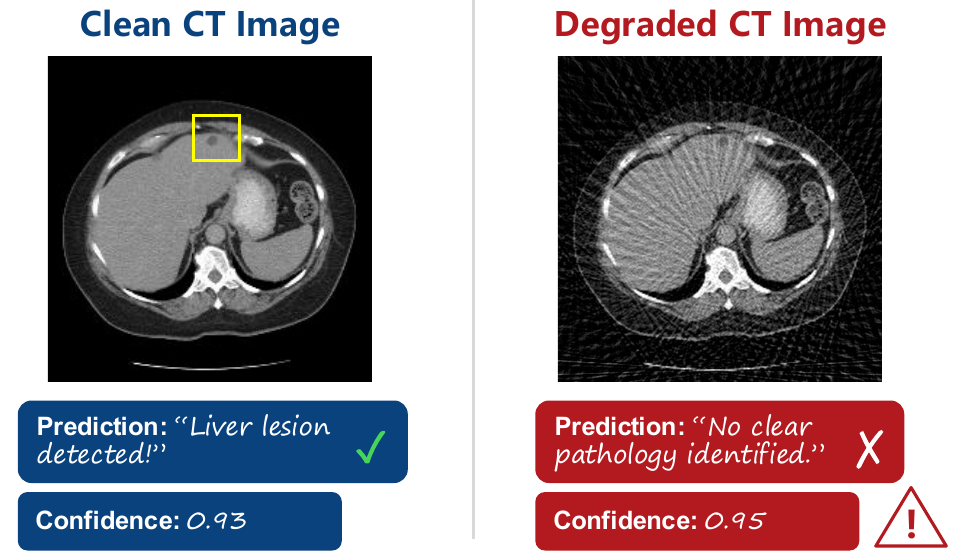}
    \caption{\textbf{Illustration of AI Dunning-Kruger Effect.} An MLLM correctly identifies a liver lesion in the clean CT image but yields erroneous predictions with similarly high confidence when the image is corrupted while the lesion is still visible.
    }
    \label{fig:overconfidence}
    % \vspace{-1em}
  \end{wrapfigure}

  This vulnerability becomes evident when we consider real-world clinical environments that are far more complex than standard benchmark datasets suggest. Medical images in nature suffer from various degradations, including noise from low-dose scanning, motion artifacts from patient movement, field inhomogeneity from aging equipment, and illumination variations in endoscopy, particularly in low-resource settings or smaller clinics where advanced imaging equipment is unavailable. As illustrated in \cref{fig:all_models_comparison}, models exhibit consistent accuracy decline across increasing degradation severity degrees. While human physicians can usually navigate these imperfections and still reach a correct diagnosis, the reliability of MLLMs in such scenarios remains largely unproven, creating a significant barrier to safe deployment in healthcare.

More alarmingly, our pilot investigations reveal a systematic pattern of metacognitive failure in MLLMs when confronted with degraded medical images. Models not only suffer accuracy drops but also exhibit a striking inability to recognize the boundaries of their own competence, maintainin inappropriately high confidence while giving erroneous predictions. As illustrated in \cref{fig:overconfidence}, an MLLM correctly detected the liver lesion in a computed tomography (CT) image when no degradation is present; however, when the same image is degraded (\eg, with sparse-view CT artifacts due to the subsampling of raw measurement data~\cite{fbpconv}), the model yields false-negative predictions while assigning similarly high confidence. This disconnect between capability and self-assessment manifests across diverse degradations and severity degrees, suggesting a fundamental deficit in model awareness rather than isolated failures.

We term this phenomenon the \textbf{AI Dunning-Kruger Effect}, analogous to the cognitive bias where individuals with limited ability overestimate their competence~\cite{dunning2011dunning}. This metacognitive blindness, \ie, the inability to ``know what it does not know,'' poses particularly severe risks in clinical deployment, as overconfident erroneous inferences may prevent the activation of necessary human oversight and mislead physicians into trusting unreliable AI recommendations.

Despite these critical vulnerabilities, existing evaluation frameworks remain inadequate for systematically assessing model performance under image quality variations. Current benchmarks suffer from two fundamental limitations: (1) \textit{Absence of large-scale multidimensional assessment under medical image degradations}. General image quality benchmarks~\cite{li2025r,qiu2025benchmarking} typically focus on natural image corruptions (\eg, Gaussian noise, blur, JPEG compression) that fail to capture medical imaging-specific degradations such as motion artifacts in magnetic resonance imaging (MRI). Although recent work~\cite{cheng2025understanding,xu2025perceive} has begun exploring medical MLLM performance under degradations, they lack the scale and comprehensiveness necessary for systematic evaluation. (2) \textit{No systematic confidence calibration analysis}. Existing medical benchmarks provide only coarse-grained overall accuracy metrics, without quantifying model confidence calibration and metacognitive awareness under degradation, which are precisely the dimensions essential for ensuring safe clinical deployment.

To systematically analyze the aforementioned phenomena and address these gaps, we propose \textbf{\benchname}, a benchmark for comprehensively evaluating medical MLLM robustness against clinically realistic image degradations. As shown in \cref{fig:framework}. Our benchmark enables both fine-grained capability evaluation under diverse medical image degradations and systematic analysis of model confidence calibration, addressing the critical dimensions missing from existing evaluation frameworks. Our contributions are threefold:

\textbf{(1) A systematic benchmark with a hierarchical evaluation framework.} We construct \benchname, which features a three-tier framework along two axes: \textit{capability} (spanning from mid-level categories through 6 clinical tasks to 30 fine-grained skills) and \textit{degradation} (covering 18 subcategories across 7 imaging modalities, with each instantiated at 3 severity degrees, each validated by expert radiologists). This hierarchical design enables structured and clinically meaningful assessment across image quality degradation types and degrees.

\textbf{(2) Quantitative evidence of the AI Dunning-Kruger Effect in medical MLLMs.} We introduce \textit{Calibration Shift}, a metric that quantifies the gap between a model's perceived confidence and its actual accuracy under degradation. Using this metric, we provide large-scale empirical evidence of the AI Dunning-Kruger Effect: medical MLLMs remain markedly overconfident even as their true capabilities deteriorate, and this overconfidence systematically widens with increasing degradation severity. This metacognitive failure reveals that current models lack the self-awareness required for safe clinical deployment.

\textbf{(3) Comprehensive evaluation across models and multidimensions.} We conduct extensive evaluation of 40 mainstream MLLMs spanning commercial MLLMs, open-source general models, and medical-specialized models. Our multidimensional analysis examines performance degradation across capability dimensions, degradation categories, and imaging modalities, providing the most comprehensive characterization of medical MLLM behavior under image quality variations to date.

\begin{table}[t]
  \centering
  \caption{\textbf{Comparison with related benchmarks.} ``--'' indicates the information is not reported in the original publication.}
  \label{tab:benchmark_comparison}
  \resizebox{\columnwidth}{!}{%
  \begin{tabular}{lccccc}
  \toprule
  \textbf{Benchmark} & \textbf{Publication} & \textbf{Total QA} & \textbf{\#Modality} & \textbf{\#Capability} & \textbf{\#Degrad.} \\
  \midrule
  MedMNIST-C~\cite{di2024medmnist} & Arxiv 2024 & 520k (train \& eval.) & 9 & 1 (classification) & 17 \\
  RobustMedCLIP~\cite{imam2025robustness} & MIUA 2025 & 137k (train \& eval.) & 5 & 1 (classification) & 7 \\
  Pathology Corruptions~\cite{zhang2022benchmarking} & MICCAI 2022 & 327k (train \& eval.) & 1 & 1 (classification) & 9 \\
  ROOD-MRI~\cite{boone2023rood} & NeuroImage 2023 & 1,475 (eval.) & 1 & 1 (segmentation) & 11 \\
  VLM Artefacts~\cite{cheng2025understanding} & npj Digit. Med. 2025 & 6,600 (eval.) & 3 & 1 (detection) & 5 \\
  RobustMed-Bench~\cite{xu2025perceive}           & Arxiv 2025  &  1,500 (eval.) & 3 & -- & 6 \\
  \midrule
  \textbf{\benchname (Ours)}   & \textbf{This paper} & \textbf{24,894} (eval.) & \textbf{7} & \textbf{30} & \textbf{18} \\
  \bottomrule
  \end{tabular}%
  }
\end{table}

\section{Related Work}

\subsection{Medical Multimodal Large Language Models and Benchmarks}

The application of Multimodal Large Language Models in medical domains has witnessed rapid advancement. To better address medical-specific challenges, a series of \textit{medical-specialized models} have emerged, including GMAI-VL-R1~\cite{su2025gmai}, UniMedVL~\cite{ning2025unimedvlunifyingmedicalmultimodal}, Hulu-Med~\cite{jiang2025hulu}, Lingshu~\cite{xu2025lingshu}, HealthGPT~\cite{lin2025healthgpt}, and MedGemma~\cite{sellergren2025medgemma}. These models are typically pre-trained or fine-tuned on large-scale medical image-text datasets to enhance domain-specific understanding.

To systematically evaluate these medical MLLMs, numerous benchmarks have been developed targeting various medical scenarios. GMAI-MMBench~\cite{gmai2024} integrates multiple medical datasets covering clinical tasks, department categorization, and perception granularity. OmniMedVQA~\cite{omnimedvqa2023} aggregates various medical VQA datasets to provide comprehensive evaluation. Modality-specific benchmarks include PathVQA~\cite{pathvqa2020} for pathology images and VQA-RAD~\cite{vqarad2018} for radiology. SLAKE~\cite{slake2021} provides bilingual medical VQA with segmentation masks. General multimodal benchmarks such as MMMU~\cite{mmmu2024} and MMMU-Pro~\cite{mmmu_pro2024} encompass multidisciplinary tasks including medical content. Recent benchmarks MedXpertQA~\cite{medxpertqa2024} for expert-level exam-style question answering.
Despite their comprehensive coverage, these benchmarks primarily evaluate performance on standard clean data without systematically considering robustness to image degradations that commonly occur in real-world clinical environments. 

\subsection{Evaluating MLLMs Under Image Degradations}
In computer vision, ImageNet-C~\cite{imagenetc2019} pioneered systematic evaluation by introducing 15 common corruption types for image classification. This framework has been extended to evaluate MLLMs under various image quality conditions~\cite{li2025res,qiu2025benchmarking}, revealing that even state-of-the-art models exhibit significant performance degradation when confronted with corrupted natural images.

Early efforts in medical imaging have explored degradation robustness within narrow scopes. MedMNIST-C~\cite{di2024medmnist} extends the MedMNIST suite with common corruptions but does not include QA evaluation or capability assessment. RobustMedCLIP~\cite{imam2025robustness} examines the robustness of medical vision-language models yet is limited to contrastive learning settings without a structured benchmark. Zhang~\etal~\cite{zhang2022benchmarking} benchmark pathology-specific corruptions for classification tasks but restrict their scope to a single modality. ROOD-MRI~\cite{boone2023rood} targets out-of-distribution robustness for MRI segmentation under 11 degradation variants, covering only one modality and one task type. While these works establish important precedents, they are largely confined to limited modalities, single tasks, or traditional vision models, leaving the robustness of MLLMs across diverse medical imaging settings substantially unexplored.

Recent benchmarks have begun examining MLLM robustness to natural image degradation, including R-Bench~\cite{li2025r} and Bench-C~\cite{sui2025benchmarking}. Some works expand the evaluation to medical MLLM behavior under image perturbations. MedQBench~\cite{liu2025medq} evaluates MLLMs' capability in assessing medical image quality. Cheng~\etal~\cite{cheng2025understanding} constructed a VLM artifacts benchmark across multiple modalities and degradation types, but evaluation is confined to disease detection without broader clinical reasoning assessment. Xu~\etal~\cite{xu2025perceive} proposes RobustMed-Bench, targeting image degradation robustness, but omits capability decomposition, severity calibration, and confidence analysis. Together, these efforts remain limited in scope, covering only a handful of modalities or degradation types without a comprehensive assessment across diverse medical tasks and capability dimensions. \cref{tab:benchmark_comparison} provides a structured comparison of \benchname against representative medical image degradation benchmarks, highlighting the gaps that motivate our work.

\section{The \benchname Benchmark}

\begin{figure*}[!t]
  % \vspace{-10pt}
  \centering
  \includegraphics[width=1.0\textwidth]{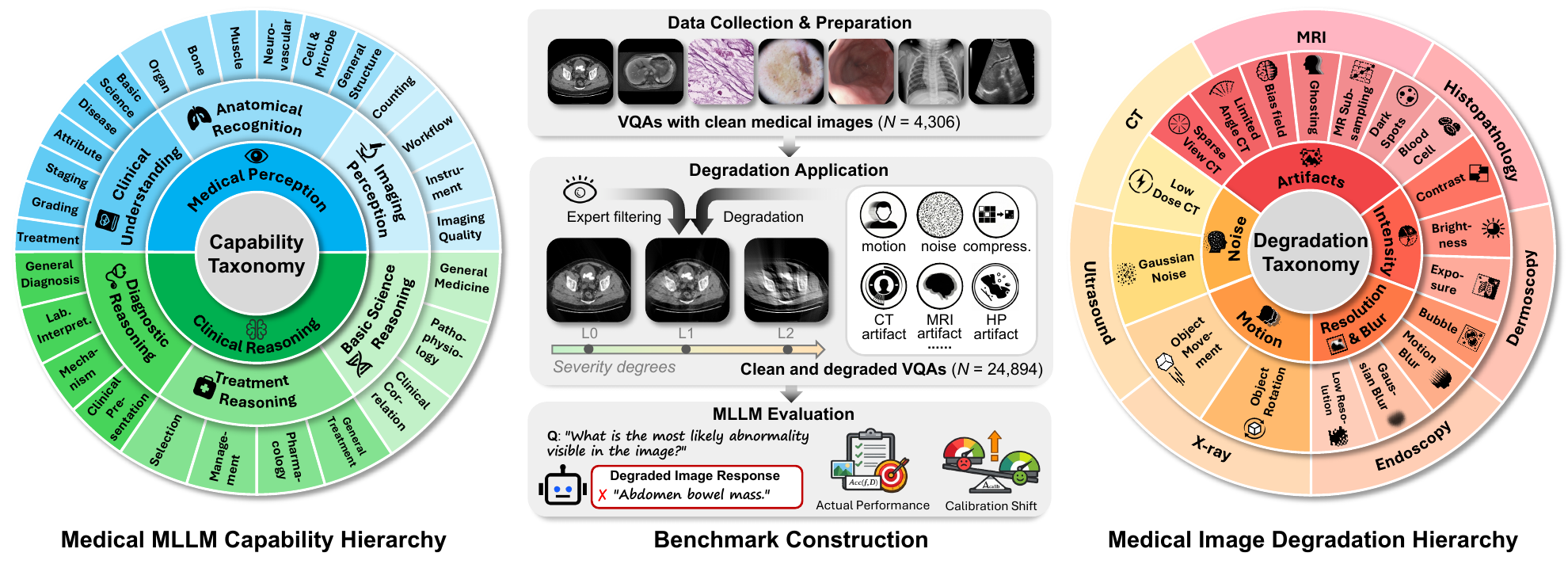}
  \caption{\textbf{Overview of the \benchname benchmark framework.} Two orthogonal hierarchies structure the evaluation: a \textit{capability hierarchy} (left) decomposing clinical competence into 30 fine-grained skills across 6 tasks, and a \textit{degradation hierarchy} (right) covering 18 degradation types across 7 modalities. The data pipeline (middle) applies each degradation at three expert-calibrated severity degrees (L0--L2).}
  \label{fig:framework}
  % \vspace{-1.0cm}
  \end{figure*}

\benchname benchmark is designed to evaluate how image quality degradation affects the diagnostic performance and metacognitive reliability of existing MLLMs. 
As illustrated in \cref{fig:framework}, the benchmark is built upon a dual-tier hierarchy that stratifies both model capabilities (Section~\ref{ssec:bench-capability}) and clinical degradation patterns (Section~\ref{ssec:bench-degradation}), supported by a human-curated data pipeline that ensures clinical realism (Section~\ref{ssec:bench-construction}) and a novel suite of metacognitive metrics designed to quantify model overconfidence under stress (Section~\ref{ssec:bench-metrics}). By doing so, we move beyond simple accuracy scores to provide multidimensional performance assessment and fine-grained capability characterization.

\subsection{Medical MLLM Capability Hierarchy}
\label{ssec:bench-capability}

To comprehensively assess MLLM clinical competence, we construct a three-tier capability hierarchy grounded in the cognitive workflow of a practising clinician (\cref{fig:framework}, left). Tasks are sourced from three top-tier medical benchmarks, namely GMAI-MMBench~\cite{gmai2024}, OmniMedVQA~\cite{omnimedvqa2023}, and MedXpertQA~\cite{medxpertqa2024}, with redundant items merged and the task structure reorganized to align with clinically meaningful capability dimensions. Full task definitions and statistics are provided in the supplementary material (Section A). The hierarchy is organized as follows:

\textbf{(1) High-level Capabilities ($\mathcal{T}^{(1)}$)}:
We partition the capability space into two fundamental categories:
\textit{medical perception} and \textit{clinical reasoning}.
The former focuses on extracting visual and structural information from medical images,
while the latter involves higher-level clinical interpretation and decision-making.
\textbf{(2) Mid-level Tasks ($\mathcal{T}^{(2)}$)}: We decompose each high-level capability into six clinically relevant dimensions that mirror the cognitive workflow of a practising clinician: \textit{anatomical recognition}, \textit{imaging perception}, \textit{clinical understanding}, \textit{basic science reasoning}, \textit{diagnostic reasoning}, and \textit{treatment reasoning} constitute the reasoning branch.
\textbf{(3) Fine-grained Skills ($\mathcal{T}^{(3)}$)}: Each mid-level task is further stratified into specific clinical skills, yielding 30 fine-grained capabilities that span the full spectrum of medical image understanding competencies. These decompositions follow the clinical diagnostic process, progressing from raw image perception to actionable clinical decision-making, and ensuring that each dimension captures a distinct and non-redundant cognitive step.

\subsection{Medical Image Degradation Hierarchy}
\label{ssec:bench-degradation}

We organize degradations into a two-level hierarchy. At the top level, five categories are grounded in clinical imaging physics: artifacts, intensity jitter, ``resolution \& blur'', motion interference, and noise. Within each category, types are further partitioned by formation mechanism into general (cross-modality, \eg, motion blur) and modality-specific corruptions (\eg, low-dose CT noise), as illustrated in the right panel of \cref{fig:framework}. Detailed definitions and statistics are provided in the supplementary material (Section A).

We define a degradation operation $\mathcal{C}_i^s$ mathematically as:
\begin{equation}
\mathcal{I}_{c} = \mathcal{C}_i^s(\mathcal{I}_0; \theta_i^s),
\end{equation}
where $\mathcal{I}_0$ denotes an original diagnostic-quality image, $i \in \{1, 2, ..., N\}$ indexes the degradation variants, and $s \in \{0, 1, 2\}$ represents the severity degree.
The severity parameters $\theta_i^s$ are calibrated independently for each degradation type: for each corruption, we present a continuous range of degradation degrees to three board-certified radiologists, who label the threshold at which diagnostic features remain intact (L1) and the threshold at which diagnosis becomes challenging yet feasible (L2). L0 retains the original clean image without degradation. 
All degradation variants are implemented using TorchIO~\cite{torchio2021}, Torch Radon~\cite{torchradon2021}, SciPy ndimage~\cite{scipy2020}, and Froodo~\cite{froodo2022}.
As all corruptions in \benchname are synthetically generated, their alignment with real-world clinical degradations will be validated in Section~\ref{app:distribution_alignment}.

\subsection{Dataset Construction}\label{ssec:bench-construction}

After cross-source deduplication based on image hash and question-text similarity, we obtain a pool of unique VQA pairs (L0) from three established benchmarks: OmniMedVQA~\cite{omnimedvqa2023}, GMAI-MMBench~\cite{gmai2024}, and MedXpertQA~\cite{medxpertqa2024}.
For each VQA pair, we randomly apply three modality-specific degradations at both severity degrees L1 and L2 on the image.

To maintain clinical validity, we implemented a human-in-the-loop filtering process in which three board-certified radiologists independently reviewed each degraded image–question pair under two criteria: (1)~the applied corruptions must not completely obliterate the diagnostic features required to answer the question, and (2)~the question must not be trivially answerable from text cues alone without reference to the image.
Samples failing either criterion were discarded.
In total, approximately 8.3\% of generated pairs were removed through this process, ensuring that every retained sample demands genuine visual reasoning under degradation.
The resulting benchmark comprises \uline{24,894 QA pairs in total}.

  \subsection{Evaluation Metrics}\label{ssec:bench-metrics}

  To assess model robustness and metacognitive capabilities under image degradation, we utilize a combination of accuracy and uncertainty metrics for quantifying both actual performance and calibration shift.
  
  \textbf{Actual Performance.} For each VQA pair in our dataset, models generate predictions through multiple-choice selection. We evaluate performance using accuracy as the primary metric:
  \begin{equation}
  \text{Acc}(f, \mathcal{D}) = \frac{1}{|\mathcal{D}|} \sum_{x \in \mathcal{D}} \mathbb{I}\bigl[\hat{y}(x) = y^\ast(x)\bigr],
  \end{equation}
  where $\mathcal{D}$ denotes the evaluation dataset, $y^\ast(x)$ represents the ground-truth answer, $\hat{y}(x)$ is the model's prediction, and $\mathbb{I}[\cdot]$ is the indicator function. 

  \textbf{Perceived Confidence.} We measure model confidence through voting-based prediction consistency. The voting distribution is $p_i(x) = \frac{1}{T} \sum_{t=1}^{T} \mathbb{I}\bigl[y^{(t)} = i\bigr]$ for $i \in \{1,\dots,K\}$, where $K$ is the number of answer choices and $y^{(t)}$ denotes the prediction in trial $t$. 
  We quantify prediction uncertainty using normalized entropy: 
  $H(x) = -\sum_{i=1}^{K} p_i(x) \log p_i(x)$, with maximum entropy $H_{\max} = \log K$ for complete uncertainty. Sample-level confidence is defined as the inverse of normalized entropy:
  \begin{equation}
      C(x) = 1 - H(x)/{\log K},
  \end{equation}
  where $C(x) \in [0,1]$, with $C(x)=1$ indicating perfect prediction consistency and $C(x) \approx 0$ for high uncertainty.

  \textbf{Calibration Shift.} Model calibration shift on dataset $\mathcal{D}$ is quantified as the gap between perceived confidence and actual accuracy:
  \begin{equation}
  \Delta_{\text{calib}}(f, \mathcal{D}) = \frac{1}{|\mathcal{D}|}\sum_{x \in \mathcal{D}} C(x) - \text{Acc}(f, \mathcal{D}).
  \end{equation}
  A positive $\Delta_\text{calib}$ indicates overconfidence (model overestimates its capability), zero represents well-calibrated confidence, and negative values indicate underconfidence. 

  We use $\Delta_\text{calib}$ to characterize two distinct forms of the AI Dunning-Kruger Effect (DKE).
\textit{Intra-Model DKE} occurs when a model’s accuracy collapses as image quality drops from L0 to L2, but the calibration shift remains stable or increases, defined as: 
\begin{equation}
  \text{Acc}(f, \mathcal{D}_{\text{L0}}) > \text{Acc}(f, \mathcal{D}_{\text{L2}}) \;\wedge\; \Delta_{\text{calib}}(f, \mathcal{D}_{\text{L0}}) \leq \Delta_{\text{calib}}(f, \mathcal{D}_{\text{L2}}),
  \end{equation}
where $\mathcal{D}_\text{L0}$ and $\mathcal{D}_\text{L2}$ denote evaluation sets at L0 and L2 severity degree respectively. This reveals a metacognitive failure where the model is ``unaware'' of its own declining capability. 
\textit{Inter-Model DKE} describes a trend that lower-performing models show higher calibration shift compared to higher-performing models:
\begin{equation}
  \text{Acc}(f_1, \mathcal{D}) < \text{Acc}(f_2, \mathcal{D}) \;\wedge\; \Delta_{\text{calib}}(f_1, \mathcal{D}) > \Delta_{\text{calib}}(f_2, \mathcal{D}).
  \end{equation}

\section{Experiments}

This section presents a comprehensive evaluation of representative MLLMs through our hierarchical benchmark. Based on the experimental results, we systematically characterize the performance of these models and provide in-depth interpretation of their strengths and limitations across diverse evaluation scenarios.
Specifically, we begin by examining overall robustness across degradation severity degrees (Section~\ref{ssec:exp-severity}), then analyze how performance varies across clinical capability dimensions (Section~\ref{ssec:exp-capability}) and degradation types (Section~\ref{ssec:exp-degradation}), characterize the overconfidence phenomenon that emerges as degradation increases (Section~\ref{ssec:exp-overconfidence}), and finally validate that the simulated degradations align with real-world clinical distributions (Section~\ref{app:distribution_alignment}).

\begin{figure*}[!t]
    \centering
    \includegraphics[width=\textwidth]{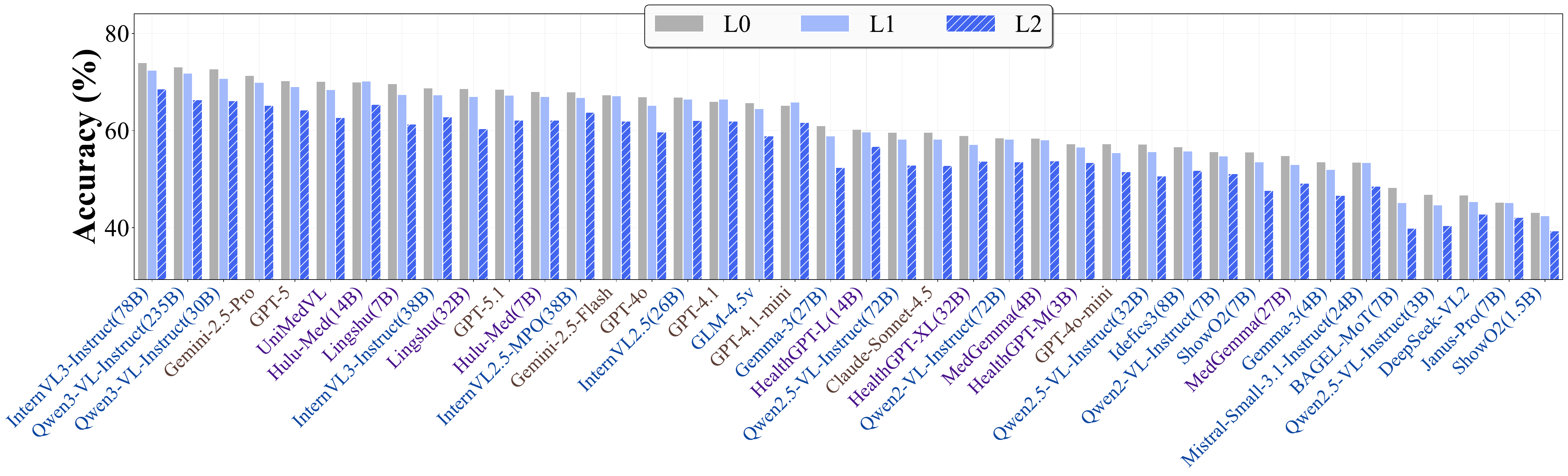}
    \caption{\textbf{Model performance across medical image quality degradation.} Accuracy of 40 MLLMs evaluated at three severity degrees (L0--L2). X-axis labels are colored by model category: blue for open-source general models, brown for commercial MLLMs, and purple for medical-specialized models.}
    \label{fig:all_models_comparison}
    \end{figure*}

\begin{table}[!ht]
  \centering
  \scriptsize
  \setlength{\tabcolsep}{7pt}
  \renewcommand{\arraystretch}{1.2}
  \caption{\textbf{Capability-level performance (L1\&L2) with degradation magnitude.} Average accuracy across the mid-level capability dimensions, computed over L1 and L2 degradations. Numbers in parentheses show performance drop from L0 to L1\&L2. ``CU'': Clinical Understanding. ``IP'': Imaging Perception. ``AR'': Anatomical Recognition. ``BS'': Basic Science. ``Diag.'': Diagnosis. ``Treat.'': Treatment.}
  \label{tab:capability_dimensions}
  \resizebox{\columnwidth}{!}{
  \begin{tabular}{lc|cccccc}
  \toprule
  & & \multicolumn{3}{c}{\textbf{Medical Perception}} & \multicolumn{3}{c}{\textbf{Clinical Reasoning}} \\
  \cmidrule(lr){3-5} \cmidrule(lr){6-8}
  \textbf{Model} & \textbf{Avg.} & \textbf{CU} & \textbf{IP} & \textbf{AR} & \textbf{BS} & \textbf{Diag.} & \textbf{Treat.} \\
  \midrule
  \noalign{\vskip -\belowrulesep}
  \multicolumn{8}{c}{\cellcolor{orange!15}\textbf{\textit{Commercial MLLMs}}} \\
  
  GPT-5 & \cellcolor{green!22}64.5{\tiny($\downarrow$3.5)} & \cellcolor{green!27}67.2{\tiny($\downarrow$4.3)} & 36.5{\tiny($\downarrow$0.9)} & \cellcolor{green!28}67.4{\tiny($\downarrow$4.4)} & \cellcolor{green!42}75.0{\tiny($\downarrow$6.0)} & \cellcolor{green!18}62.1{\tiny($\downarrow$2.9)} & \cellcolor{green!50}78.8{\tiny($\downarrow$2.3)} \\
  Gemini-2.5-Pro & \cellcolor{green!10}58.2{\tiny($\downarrow$3.8)} & \cellcolor{green!27}66.9{\tiny($\downarrow$3.4)} & 30.0{\tiny($\downarrow$6.5)} & \cellcolor{green!37}72.0{\tiny($\downarrow$5.7)} & \cellcolor{green!15}60.7{\tiny($\downarrow$8.3)} & \cellcolor{green!14}60.1{\tiny($\downarrow$1.3)} & \cellcolor{green!13}59.6{\tiny($\uparrow$2.6)} \\
  Gemini-2.5-Flash & 50.7{\tiny($\downarrow$2.9)} & \cellcolor{green!20}63.6{\tiny($\downarrow$4.8)} & \cellcolor{red!1}25.5{\tiny($\downarrow$1.2)} & \cellcolor{green!35}71.2{\tiny($\uparrow$0.1)} & \cellcolor{green!12}58.9{\tiny($\downarrow$5.7)} & 38.9{\tiny($\downarrow$3.2)} & 46.1{\tiny($\downarrow$2.8)} \\
  GPT-5.1 & 49.6{\tiny($\downarrow$5.9)} & \cellcolor{green!25}65.9{\tiny($\downarrow$3.3)} & 30.5{\tiny($\downarrow$6.0)} & \cellcolor{green!25}65.8{\tiny($\downarrow$4.2)} & 50.0{\tiny($\downarrow$9.5)} & 47.0{\tiny($\downarrow$8.5)} & 38.5{\tiny($\downarrow$3.9)} \\
  GPT-4.1 & 48.9{\tiny($\downarrow$2.3)} & \cellcolor{green!26}66.5{\tiny($\downarrow$2.8)} & 32.5{\tiny($\downarrow$1.9)} & \cellcolor{green!21}63.8{\tiny($\downarrow$1.1)} & 50.0{\tiny($\downarrow$4.1)} & 38.4{\tiny($\downarrow$2.1)} & 42.3{\tiny($\downarrow$1.8)} \\
  
  GPT-4o & 47.5{\tiny($\downarrow$3.2)} & \cellcolor{green!23}64.8{\tiny($\downarrow$5.6)} & 32.0{\tiny($\downarrow$2.5)} & \cellcolor{green!17}61.9{\tiny($\downarrow$4.3)} & 42.9{\tiny($\uparrow$0.2)} & 37.4{\tiny($\downarrow$3.9)} & 46.1{\tiny($\downarrow$3.0)} \\
  GPT-4.1-mini & 47.0{\tiny($\downarrow$1.6)} & \cellcolor{green!25}65.9{\tiny($\downarrow$1.9)} & 34.0{\tiny($\downarrow$2.0)} & \cellcolor{green!20}63.5{\tiny($\downarrow$1.8)} & 33.9{\tiny($\downarrow$1.2)} & 44.4{\tiny($\downarrow$2.7)} & 40.4{\tiny($\downarrow$0.1)} \\
  Claude-Sonnet-4.5 & 45.1{\tiny($\downarrow$3.1)} & \cellcolor{green!19}62.8{\tiny($\downarrow$3.8)} & \cellcolor{red!8}22.0{\tiny($\downarrow$4.5)} & 44.3{\tiny($\downarrow$6.3)} & 51.8{\tiny($\uparrow$2.3)} & 35.9{\tiny($\downarrow$4.6)} & \cellcolor{green!2}53.9{\tiny($\downarrow$1.4)} \\
  GPT-4o-mini & 31.6{\tiny($\downarrow$2.9)} & \cellcolor{green!9}57.3{\tiny($\downarrow$3.8)} & \cellcolor{red!3}24.5{\tiny($\downarrow$6.7)} & 50.4{\tiny($\downarrow$5.7)} & \cellcolor{red!26}12.5{\tiny($\downarrow$0.1)} & 29.3{\tiny($\downarrow$0.2)} & \cellcolor{red!20}15.4{\tiny($\downarrow$0.9)} \\
  \rowcolor{orange!5}\textit{Avg.~Commercial MLLMs} & 49.2{\tiny($\downarrow$3.2)} & 64.5{\tiny($\downarrow$3.7)} & 29.7{\tiny($\downarrow$3.6)} & 62.3{\tiny($\downarrow$3.7)} & 48.4{\tiny($\downarrow$3.6)} & 43.7{\tiny($\downarrow$3.3)} & 46.8{\tiny($\downarrow$1.5)} \\
  
  \midrule
  \noalign{\vskip -\belowrulesep}
  \multicolumn{8}{c}{\cellcolor{cyan!15}\textbf{\textit{Open-Source General Models}}} \\
  
  Qwen3-VL-Instruct(235B) & 43.8{\tiny($\downarrow$4.2)} & \cellcolor{green!38}73.0{\tiny($\downarrow$3.7)} & 40.5{\tiny($\downarrow$8.1)} & \cellcolor{green!27}66.8{\tiny($\downarrow$4.9)} & 37.5{\tiny($\downarrow$2.3)} & 29.8{\tiny($\downarrow$4.9)} & \cellcolor{red!20}15.4{\tiny($\downarrow$1.5)} \\
  InternVL3-Instruct(78B) & 39.3{\tiny($\downarrow$3.8)} & \cellcolor{green!44}75.9{\tiny($\downarrow$3.1)} & 35.0{\tiny($\uparrow$0.3)} & \cellcolor{green!26}66.3{\tiny($\downarrow$3.1)} & \cellcolor{red!9}21.4{\tiny($\downarrow$9.1)} & 29.3{\tiny($\downarrow$7.0)} & \cellcolor{red!35}7.7{\tiny($\downarrow$0.6)} \\
  GLM-4.5v & 38.5{\tiny($\downarrow$3.5)} & \cellcolor{green!26}66.4{\tiny($\downarrow$3.6)} & \cellcolor{red!1}25.5{\tiny($\downarrow$1.7)} & \cellcolor{green!9}57.7{\tiny($\downarrow$6.3)} & 37.5{\tiny($\downarrow$5.1)} & 32.3{\tiny($\uparrow$0.1)} & \cellcolor{red!28}11.5{\tiny($\downarrow$4.3)} \\
  Qwen3-VL-Instruct(30B) & 38.1{\tiny($\downarrow$4.1)} & \cellcolor{green!38}73.0{\tiny($\downarrow$2.6)} & \cellcolor{red!1}25.5{\tiny($\downarrow$4.7)} & \cellcolor{green!25}66.1{\tiny($\downarrow$8.1)} & \cellcolor{red!22}14.3{\tiny($\downarrow$9.1)} & 30.3{\tiny($\uparrow$1.8)} & \cellcolor{red!13}19.2{\tiny($\downarrow$2.1)} \\
  InternVL2.5-MPO(38B) & 36.5{\tiny($\downarrow$1.6)} & \cellcolor{green!38}72.6{\tiny($\downarrow$2.3)} & 30.0{\tiny($\downarrow$0.1)} & \cellcolor{green!8}57.0{\tiny($\downarrow$4.1)} & \cellcolor{red!2}25.0{\tiny($\downarrow$1.8)} & \cellcolor{red!6}22.7{\tiny($\downarrow$0.5)} & \cellcolor{red!28}11.5{\tiny($\downarrow$0.8)} \\
  Mistral-Small-3.1-Instruct(24B) & 33.9{\tiny($\downarrow$2.5)} & \cellcolor{green!7}56.3{\tiny($\downarrow$2.9)} & \cellcolor{red!2}25.0{\tiny($\downarrow$0.3)} & 43.9{\tiny($\downarrow$2.3)} & \cellcolor{red!2}25.0{\tiny($\downarrow$0.7)} & 33.8{\tiny($\downarrow$1.3)} & \cellcolor{red!13}19.2{\tiny($\downarrow$7.7)} \\
  InternVL3-Instruct(38B) & 33.4{\tiny($\downarrow$4.0)} & \cellcolor{green!36}71.7{\tiny($\downarrow$3.0)} & \cellcolor{red!4}24.0{\tiny($\downarrow$8.0)} & \cellcolor{green!10}58.3{\tiny($\downarrow$5.8)} & \cellcolor{red!29}10.7{\tiny($\downarrow$4.5)} & 31.8{\tiny($\downarrow$2.6)} & \cellcolor{red!42}3.9{\tiny($\downarrow$0.3)} \\
  InternVL2.5(26B) & 33.0{\tiny($\downarrow$2.4)} & \cellcolor{green!36}71.8{\tiny($\downarrow$2.8)} & 27.5{\tiny($\downarrow$6.4)} & \cellcolor{green!6}55.7{\tiny($\downarrow$4.5)} & \cellcolor{red!26}12.5{\tiny($\downarrow$0.4)} & \cellcolor{red!6}22.7{\tiny($\downarrow$0.1)} & \cellcolor{red!35}7.7{\tiny($\uparrow$0.1)} \\
  Qwen2.5-VL-Instruct(72B) & 31.8{\tiny($\downarrow$4.0)} & \cellcolor{green!20}63.4{\tiny($\downarrow$4.3)} & 28.5{\tiny($\downarrow$3.7)} & 44.5{\tiny($\downarrow$5.7)} & \cellcolor{red!12}19.6{\tiny($\downarrow$2.3)} & 28.8{\tiny($\downarrow$5.9)} & \cellcolor{red!39}5.8{\tiny($\downarrow$1.9)} \\
  Gemma-3(27B) & 31.5{\tiny($\downarrow$3.2)} & \cellcolor{green!11}58.7{\tiny($\downarrow$4.9)} & \cellcolor{red!6}23.0{\tiny($\downarrow$0.1)} & \cellcolor{green!4}55.2{\tiny($\downarrow$6.1)} & \cellcolor{red!22}14.3{\tiny($\downarrow$3.6)} & \cellcolor{red!14}18.7{\tiny($\downarrow$4.6)} & \cellcolor{red!13}19.2{\tiny($\uparrow$0.1)} \\
  Gemma-3(4B) & 31.5{\tiny($\downarrow$4.3)} & \cellcolor{green!3}54.4{\tiny($\downarrow$2.6)} & 32.5{\tiny($\downarrow$0.1)} & 43.2{\tiny($\downarrow$7.2)} & \cellcolor{red!16}17.9{\tiny($\downarrow$15.0)} & \cellcolor{red!23}14.1{\tiny($\downarrow$1.2)} & 26.9{\tiny($\uparrow$0.1)} \\
  
  Qwen2-VL-Instruct(7B) & 31.4{\tiny($\downarrow$2.5)} & \cellcolor{green!17}61.7{\tiny($\downarrow$2.8)} & 30.5{\tiny($\downarrow$1.6)} & 39.7{\tiny($\downarrow$3.0)} & \cellcolor{red!5}23.2{\tiny($\uparrow$1.5)} & \cellcolor{red!12}19.7{\tiny($\downarrow$3.3)} & \cellcolor{red!24}13.5{\tiny($\downarrow$5.8)} \\
  Idefics3(8B) & 31.0{\tiny($\downarrow$1.5)} & \cellcolor{green!14}60.0{\tiny($\downarrow$3.0)} & \cellcolor{red!2}25.0{\tiny($\downarrow$0.8)} & 46.2{\tiny($\downarrow$3.6)} & \cellcolor{red!16}17.9{\tiny($\downarrow$1.1)} & \cellcolor{red!1}25.2{\tiny($\downarrow$0.5)} & \cellcolor{red!28}11.5{\tiny($\uparrow$0.2)} \\
  Qwen2.5-VL-Instruct(32B) & 29.9{\tiny($\downarrow$1.4)} & \cellcolor{green!17}61.9{\tiny($\downarrow$4.4)} & 27.5{\tiny($\downarrow$0.1)} & 39.7{\tiny($\downarrow$4.0)} & \cellcolor{red!16}17.9{\tiny($\uparrow$0.1)} & 32.3{\tiny($\downarrow$0.1)} & \cellcolor{red!50}0.0{\tiny($\downarrow$0.0)} \\
  Qwen2-VL-Instruct(72B) & 27.3{\tiny($\downarrow$3.4)} & \cellcolor{green!19}62.8{\tiny($\downarrow$3.3)} & \cellcolor{red!4}24.0{\tiny($\downarrow$1.6)} & 47.8{\tiny($\downarrow$3.1)} & \cellcolor{red!39}5.4{\tiny($\downarrow$0.3)} & \cellcolor{red!11}20.2{\tiny($\downarrow$0.9)} & \cellcolor{red!42}3.9{\tiny($\downarrow$11.5)} \\
  Janus-Pro(7B) & 27.1{\tiny($\downarrow$1.8)} & 50.7{\tiny($\downarrow$1.5)} & 37.0{\tiny($\downarrow$4.2)} & 32.7{\tiny($\downarrow$2.7)} & 30.4{\tiny($\downarrow$1.5)} & \cellcolor{red!27}11.6{\tiny($\downarrow$1.0)} & \cellcolor{red!50}0.0{\tiny($\downarrow$0.0)} \\
  ShowO2(7B) & 26.8{\tiny($\downarrow$3.0)} & \cellcolor{green!11}58.6{\tiny($\downarrow$4.0)} & \cellcolor{red!2}25.0{\tiny($\downarrow$0.1)} & 39.2{\tiny($\downarrow$8.2)} & \cellcolor{red!36}7.1{\tiny($\uparrow$0.1)} & \cellcolor{red!9}21.2{\tiny($\downarrow$0.1)} & \cellcolor{red!31}9.6{\tiny($\downarrow$5.7)} \\
  ShowO2(1.5B) & 26.6{\tiny($\downarrow$2.6)} & 46.3{\tiny($\downarrow$2.6)} & \cellcolor{red!4}24.0{\tiny($\downarrow$5.3)} & 33.2{\tiny($\downarrow$1.2)} & \cellcolor{red!9}21.4{\tiny($\downarrow$4.5)} & \cellcolor{red!20}15.7{\tiny($\downarrow$0.5)} & \cellcolor{red!13}19.2{\tiny($\downarrow$1.8)} \\
  Qwen2.5-VL-Instruct(3B) & \cellcolor{red!1}25.4{\tiny($\downarrow$3.4)} & 48.7{\tiny($\downarrow$5.0)} & 28.0{\tiny($\downarrow$3.1)} & 33.6{\tiny($\downarrow$4.5)} & 26.8{\tiny($\downarrow$6.8)} & \cellcolor{red!21}15.2{\tiny($\downarrow$0.8)} & \cellcolor{red!50}0.0{\tiny($\downarrow$0.0)} \\
  BAGEL-MoT(7B) & \cellcolor{red!3}24.7{\tiny($\downarrow$2.9)} & 47.7{\tiny($\downarrow$5.6)} & 32.5{\tiny($\downarrow$2.4)} & 35.4{\tiny($\downarrow$7.9)} & \cellcolor{red!26}12.5{\tiny($\downarrow$0.5)} & \cellcolor{red!19}16.2{\tiny($\downarrow$0.8)} & \cellcolor{red!42}3.9{\tiny($\uparrow$0.1)} \\
  DeepSeek-VL2 & \cellcolor{red!5}23.2{\tiny($\downarrow$1.7)} & 47.4{\tiny($\downarrow$1.9)} & 29.5{\tiny($\downarrow$2.8)} & 42.0{\tiny($\downarrow$3.9)} & \cellcolor{red!36}7.1{\tiny($\downarrow$0.4)} & \cellcolor{red!47}1.5{\tiny($\uparrow$0.3)} & \cellcolor{red!28}11.5{\tiny($\downarrow$1.2)} \\
  \rowcolor{cyan!5}\textit{Avg.~Open-Source Models} & 31.6{\tiny($\downarrow$2.9)} & 61.1{\tiny($\downarrow$3.3)} & 28.6{\tiny($\downarrow$2.6)} & 47.8{\tiny($\downarrow$4.8)} & 19.3{\tiny($\downarrow$3.2)} & 22.5{\tiny($\downarrow$1.6)} & 10.5{\tiny($\downarrow$2.1)} \\
  
  \midrule
  \noalign{\vskip -\belowrulesep}
  \multicolumn{8}{c}{\cellcolor{blue!15}\textbf{\textit{Medical-specialized Models}}} \\
  
  UniMedVL & 42.5{\tiny($\downarrow$4.0)} & \cellcolor{green!31}69.2{\tiny($\downarrow$3.6)} & 40.0{\tiny($\downarrow$3.7)} & \cellcolor{green!20}63.1{\tiny($\downarrow$5.6)} & \cellcolor{red!2}25.0{\tiny($\downarrow$1.6)} & 32.8{\tiny($\downarrow$7.6)} & \cellcolor{red!2}25.0{\tiny($\downarrow$1.9)} \\
  Hulu-Med(7B) & 38.8{\tiny($\downarrow$2.4)} & \cellcolor{green!35}71.0{\tiny($\downarrow$4.2)} & 31.0{\tiny($\uparrow$1.0)} & \cellcolor{green!9}57.5{\tiny($\downarrow$5.0)} & 28.6{\tiny($\downarrow$3.4)} & 27.3{\tiny($\downarrow$0.9)} & \cellcolor{red!17}17.3{\tiny($\downarrow$1.9)} \\
  Hulu-Med(14B) & 38.4{\tiny($\downarrow$1.5)} & \cellcolor{green!38}72.8{\tiny($\downarrow$2.1)} & 34.0{\tiny($\downarrow$2.3)} & \cellcolor{green!21}64.0{\tiny($\downarrow$4.6)} & 26.8{\tiny($\downarrow$0.1)} & \cellcolor{red!1}25.2{\tiny($\downarrow$0.1)} & \cellcolor{red!35}7.7{\tiny($\uparrow$0.1)} \\
  Lingshu(32B) & 36.1{\tiny($\downarrow$3.9)} & \cellcolor{green!33}70.4{\tiny($\downarrow$3.8)} & 27.5{\tiny($\uparrow$0.5)} & \cellcolor{green!6}56.2{\tiny($\downarrow$8.7)} & 26.8{\tiny($\downarrow$8.3)} & 27.8{\tiny($\downarrow$2.9)} & \cellcolor{red!35}7.7{\tiny($\downarrow$0.5)} \\
  Lingshu(7B) & 35.8{\tiny($\downarrow$5.1)} & \cellcolor{green!36}71.7{\tiny($\downarrow$4.4)} & \cellcolor{red!5}23.5{\tiny($\downarrow$8.6)} & \cellcolor{green!7}56.3{\tiny($\downarrow$8.1)} & \cellcolor{red!5}23.2{\tiny($\downarrow$2.1)} & \cellcolor{red!10}20.7{\tiny($\downarrow$6.3)} & \cellcolor{red!13}19.2{\tiny($\downarrow$1.4)} \\
  
  HealthGPT-L(14B) & 35.2{\tiny($\downarrow$2.1)} & \cellcolor{green!26}66.6{\tiny($\downarrow$1.5)} & 33.0{\tiny($\downarrow$4.6)} & 45.3{\tiny($\downarrow$4.3)} & \cellcolor{red!16}17.9{\tiny($\downarrow$4.2)} & 40.9{\tiny($\uparrow$2.8)} & \cellcolor{red!35}7.7{\tiny($\downarrow$0.6)} \\
  MedGemma(4B) & 35.0{\tiny($\downarrow$1.9)} & \cellcolor{green!18}62.1{\tiny($\downarrow$2.9)} & 30.0{\tiny($\downarrow$1.9)} & 48.0{\tiny($\downarrow$2.5)} & \cellcolor{red!16}17.9{\tiny($\uparrow$2.3)} & 26.8{\tiny($\downarrow$0.6)} & \cellcolor{red!2}25.0{\tiny($\downarrow$5.8)} \\
  HealthGPT-M(3B) & 32.6{\tiny($\downarrow$1.9)} & \cellcolor{green!21}63.7{\tiny($\downarrow$2.6)} & 28.5{\tiny($\downarrow$0.8)} & 42.0{\tiny($\downarrow$1.9)} & \cellcolor{red!22}14.3{\tiny($\downarrow$3.7)} & 27.8{\tiny($\downarrow$0.7)} & \cellcolor{red!13}19.2{\tiny($\downarrow$1.7)} \\
  MedGemma(27B) & 31.4{\tiny($\downarrow$2.4)} & \cellcolor{green!12}59.3{\tiny($\downarrow$4.1)} & 26.5{\tiny($\downarrow$0.1)} & 38.0{\tiny($\downarrow$3.5)} & \cellcolor{red!29}10.7{\tiny($\uparrow$0.1)} & 42.4{\tiny($\downarrow$5.6)} & \cellcolor{red!28}11.5{\tiny($\downarrow$1.1)} \\
  HealthGPT-XL(32B) & 30.5{\tiny($\downarrow$3.2)} & \cellcolor{green!16}61.4{\tiny($\downarrow$3.9)} & 33.0{\tiny($\downarrow$0.3)} & 48.5{\tiny($\downarrow$2.6)} & \cellcolor{red!16}17.9{\tiny($\downarrow$9.1)} & \cellcolor{red!22}14.7{\tiny($\downarrow$2.6)} & \cellcolor{red!35}7.7{\tiny($\downarrow$0.5)} \\
  \rowcolor{blue!5}\textit{Avg.~Medical-Specialized} & 35.6{\tiny($\downarrow$2.8)} & 66.8{\tiny($\downarrow$3.3)} & 30.7{\tiny($\downarrow$2.1)} & 51.9{\tiny($\downarrow$4.7)} & 20.9{\tiny($\downarrow$3.0)} & 28.6{\tiny($\downarrow$2.4)} & 14.8{\tiny($\downarrow$1.5)} \\
  \midrule
  \textbf{Avg. Performance (All Models)} & \textbf{36.6}{\tiny($\downarrow$3.0)} & \textbf{63.3}{\tiny($\downarrow$3.4)} & \textbf{29.4}{\tiny($\downarrow$2.7)} & \textbf{52.1}{\tiny($\downarrow$4.5)} & \textbf{26.3}{\tiny($\downarrow$3.2)} & \textbf{28.8}{\tiny($\downarrow$2.2)} & \textbf{19.8}{\tiny($\downarrow$1.8)} \\
  \bottomrule
  \end{tabular}
  }
  \end{table}

\subsection{Experimental Setup}
\label{ssec:exp-setup}

\begin{table}[!ht]
  \centering
  \scriptsize
  \setlength{\tabcolsep}{7pt}
  \renewcommand{\arraystretch}{1.2}
  \caption{\textbf{Degradation-category performance (L1\&L2) with degradation magnitude.} Average accuracy across five degradation categories, computed over L1 and L2 degradations. Numbers in parentheses show performance drop from L0 to L1\&L2. ``R\&B'': resolution and blur issues. }
  \label{tab:degradation_categories}
  \resizebox{\columnwidth}{!}{
\begin{tabular}{lc|ccccc}
\toprule
\textbf{Model} & \textbf{Avg.} & \textbf{Intensity} & \textbf{Noise} & \textbf{R\&B} & \textbf{Artifacts} & \textbf{Motion} \\
\midrule
\noalign{\vskip -\belowrulesep}
\multicolumn{7}{c}{\cellcolor{orange!15}\textbf{\textit{Commercial MLLMs}}} \\

Gemini-2.5-Pro & \cellcolor{green!20}68.2{\tiny($\downarrow$3.7)} & \cellcolor{green!32}71.1{\tiny($\downarrow$0.9)} & \cellcolor{green!25}69.2{\tiny($\downarrow$2.3)} & \cellcolor{green!18}67.2{\tiny($\downarrow$2.0)} & \cellcolor{green!11}65.1{\tiny($\downarrow$5.3)} & \cellcolor{green!23}68.6{\tiny($\downarrow$8.0)} \\
GPT-5 & \cellcolor{green!17}67.5{\tiny($\downarrow$4.2)} & \cellcolor{green!36}72.2{\tiny($\downarrow$4.0)} & \cellcolor{green!15}66.2{\tiny($\downarrow$3.2)} & \cellcolor{green!19}67.5{\tiny($\downarrow$2.5)} & \cellcolor{green!4}63.1{\tiny($\downarrow$6.4)} & \cellcolor{green!22}68.4{\tiny($\downarrow$4.8)} \\
GPT-5.1 & \cellcolor{green!10}65.7{\tiny($\downarrow$4.2)} & \cellcolor{green!31}70.9{\tiny($\downarrow$2.5)} & \cellcolor{green!8}64.2{\tiny($\downarrow$4.8)} & \cellcolor{green!13}65.7{\tiny($\downarrow$4.2)} & 60.5{\tiny($\downarrow$6.2)} & \cellcolor{green!18}67.2{\tiny($\downarrow$3.4)} \\
Gemini-2.5-Flash & \cellcolor{green!9}65.4{\tiny($\downarrow$2.5)} & \cellcolor{green!19}67.5{\tiny($\downarrow$0.6)} & \cellcolor{green!11}65.1{\tiny($\downarrow$1.8)} & \cellcolor{green!10}64.8{\tiny($\downarrow$1.0)} & 60.7{\tiny($\downarrow$5.0)} & \cellcolor{green!23}68.7{\tiny($\downarrow$4.0)} \\
GPT-4.1 & \cellcolor{green!8}65.0{\tiny($\downarrow$2.1)} & \cellcolor{green!26}69.4{\tiny($\downarrow$3.7)} & \cellcolor{green!3}62.7{\tiny($\downarrow$0.8)} & \cellcolor{green!11}65.1{\tiny($\downarrow$2.4)} & 60.6{\tiny($\downarrow$3.2)} & \cellcolor{green!19}67.4{\tiny($\downarrow$0.6)} \\
GPT-4.1-mini & \cellcolor{green!6}64.4{\tiny($\downarrow$1.6)} & \cellcolor{green!18}67.1{\tiny($\downarrow$1.6)} & \cellcolor{green!5}63.3{\tiny($\uparrow$0.8)} & \cellcolor{green!8}64.4{\tiny($\downarrow$1.4)} & 60.5{\tiny($\downarrow$3.1)} & \cellcolor{green!17}66.8{\tiny($\downarrow$2.5)} \\
GPT-4o & \cellcolor{green!2}63.4{\tiny($\downarrow$4.5)} & \cellcolor{green!24}68.8{\tiny($\downarrow$1.3)} & 61.8{\tiny($\downarrow$5.5)} & \cellcolor{green!4}63.2{\tiny($\downarrow$4.6)} & 58.3{\tiny($\downarrow$6.7)} & \cellcolor{green!11}65.0{\tiny($\downarrow$4.2)} \\
GPT-4o-mini & 54.7{\tiny($\downarrow$4.5)} & \cellcolor{green!1}62.3{\tiny($\downarrow$2.7)} & 53.9{\tiny($\downarrow$4.3)} & 55.1{\tiny($\downarrow$3.2)} & 47.8{\tiny($\downarrow$6.0)} & 54.2{\tiny($\downarrow$6.1)} \\
Claude-Sonnet-4.5 & 55.7{\tiny($\downarrow$5.1)} & 59.1{\tiny($\downarrow$2.7)} & 53.3{\tiny($\downarrow$5.6)} & 56.6{\tiny($\downarrow$4.5)} & 55.2{\tiny($\downarrow$7.0)} & 54.3{\tiny($\downarrow$5.5)} \\
\rowcolor{orange!5}\textit{Avg.~Commercial MLLMs} & 63.3{\tiny($\downarrow$3.6)} & 67.6{\tiny($\downarrow$2.2)} & 62.2{\tiny($\downarrow$3.1)} & 63.3{\tiny($\downarrow$2.9)} & 59.1{\tiny($\downarrow$5.4)} & 64.5{\tiny($\downarrow$4.3)} \\

\midrule
\noalign{\vskip -\belowrulesep}
\multicolumn{7}{c}{\cellcolor{cyan!15}\textbf{\textit{Open-Source General Models}}} \\

InternVL3-Instruct(78B) & \cellcolor{green!30}71.4{\tiny($\downarrow$3.2)} & \cellcolor{green!46}75.1{\tiny($\downarrow$3.6)} & \cellcolor{green!26}69.3{\tiny($\downarrow$2.8)} & \cellcolor{green!37}72.7{\tiny($\downarrow$2.0)} & \cellcolor{green!8}64.3{\tiny($\downarrow$5.2)} & \cellcolor{green!47}75.6{\tiny($\downarrow$2.5)} \\
Qwen3-VL-Instruct(235B) & \cellcolor{green!25}70.0{\tiny($\downarrow$4.1)} & \cellcolor{green!47}75.3{\tiny($\downarrow$2.9)} & \cellcolor{green!27}69.8{\tiny($\downarrow$2.7)} & \cellcolor{green!31}70.7{\tiny($\downarrow$2.2)} & \cellcolor{green!5}63.3{\tiny($\downarrow$6.2)} & \cellcolor{green!32}71.1{\tiny($\downarrow$6.5)} \\
Qwen3-VL-Instruct(30B) & \cellcolor{green!23}69.4{\tiny($\downarrow$4.2)} & \cellcolor{green!50}76.2{\tiny($\downarrow$1.0)} & \cellcolor{green!21}68.0{\tiny($\downarrow$4.5)} & \cellcolor{green!33}71.3{\tiny($\downarrow$2.2)} & 61.4{\tiny($\downarrow$7.1)} & \cellcolor{green!29}70.3{\tiny($\downarrow$6.4)} \\
InternVL2.5-MPO(38B) & \cellcolor{green!12}66.0{\tiny($\downarrow$2.6)} & \cellcolor{green!28}70.1{\tiny($\downarrow$1.4)} & \cellcolor{green!8}64.2{\tiny($\downarrow$1.7)} & \cellcolor{green!16}66.5{\tiny($\downarrow$2.4)} & \cellcolor{green!0}61.9{\tiny($\downarrow$4.4)} & \cellcolor{green!19}67.4{\tiny($\downarrow$3.1)} \\
InternVL3-Instruct(38B) & \cellcolor{green!11}65.8{\tiny($\downarrow$3.8)} & \cellcolor{green!30}70.5{\tiny($\downarrow$1.6)} & \cellcolor{green!7}64.0{\tiny($\downarrow$4.3)} & \cellcolor{green!19}67.4{\tiny($\downarrow$2.7)} & 59.9{\tiny($\downarrow$6.2)} & \cellcolor{green!18}67.2{\tiny($\downarrow$4.1)} \\
InternVL2.5(26B) & \cellcolor{green!8}65.2{\tiny($\downarrow$3.5)} & \cellcolor{green!41}73.6{\tiny($\downarrow$3.2)} & \cellcolor{green!2}62.6{\tiny($\downarrow$1.9)} & \cellcolor{green!15}66.3{\tiny($\downarrow$3.0)} & 59.9{\tiny($\downarrow$4.1)} & \cellcolor{green!6}63.7{\tiny($\downarrow$5.3)} \\
GLM-4.5v & 62.4{\tiny($\downarrow$4.6)} & \cellcolor{green!12}65.5{\tiny($\downarrow$1.7)} & \cellcolor{green!7}63.9{\tiny($\downarrow$0.6)} & \cellcolor{green!2}62.6{\tiny($\downarrow$3.8)} & 57.2{\tiny($\downarrow$7.4)} & \cellcolor{green!4}63.0{\tiny($\downarrow$9.4)} \\
Qwen2-VL-Instruct(72B) & 56.2{\tiny($\downarrow$3.5)} & 58.5{\tiny($\downarrow$3.5)} & 54.8{\tiny($\downarrow$2.4)} & 57.5{\tiny($\downarrow$2.5)} & 53.2{\tiny($\downarrow$5.4)} & 57.1{\tiny($\downarrow$3.5)} \\
Gemma-3(27B) & 56.5{\tiny($\downarrow$4.4)} & 58.2{\tiny($\downarrow$1.7)} & 58.7{\tiny($\downarrow$1.8)} & 55.2{\tiny($\downarrow$5.7)} & 51.4{\tiny($\downarrow$7.0)} & 59.1{\tiny($\downarrow$5.6)} \\
Qwen2.5-VL-Instruct(72B) & 55.8{\tiny($\downarrow$4.8)} & 57.6{\tiny($\downarrow$3.4)} & 53.0{\tiny($\downarrow$5.1)} & 57.2{\tiny($\downarrow$3.7)} & 53.6{\tiny($\downarrow$7.5)} & 57.5{\tiny($\downarrow$4.1)} \\
Idefics3(8B) & 55.3{\tiny($\downarrow$2.9)} & \cellcolor{green!13}65.7{\tiny($\downarrow$0.9)} & 50.5{\tiny($\downarrow$4.7)} & 55.7{\tiny($\downarrow$3.1)} & 48.0{\tiny($\downarrow$2.3)} & 56.6{\tiny($\downarrow$3.6)} \\
Qwen2.5-VL-Instruct(32B) & 53.0{\tiny($\downarrow$4.6)} & 50.6{\tiny($\downarrow$4.3)} & 51.8{\tiny($\downarrow$4.2)} & 54.2{\tiny($\downarrow$2.5)} & 52.3{\tiny($\downarrow$7.2)} & 56.0{\tiny($\downarrow$4.7)} \\
Qwen2-VL-Instruct(7B) & 53.0{\tiny($\downarrow$3.0)} & 53.7{\tiny($\downarrow$1.2)} & 52.6{\tiny($\downarrow$2.7)} & 53.1{\tiny($\downarrow$2.2)} & 53.0{\tiny($\downarrow$4.5)} & 52.5{\tiny($\downarrow$4.3)} \\
Mistral-Small-3.1-Instruct(24B) & 51.3{\tiny($\downarrow$2.6)} & 52.4{\tiny($\downarrow$0.2)} & 49.0{\tiny($\downarrow$2.8)} & 51.8{\tiny($\downarrow$0.5)} & 49.5{\tiny($\downarrow$5.5)} & 53.7{\tiny($\downarrow$3.8)} \\
ShowO2(7B) & 52.0{\tiny($\downarrow$4.9)} & \cellcolor{green!1}62.3{\tiny($\downarrow$1.6)} & \cellcolor{red!1}47.0{\tiny($\downarrow$6.3)} & 52.8{\tiny($\downarrow$5.9)} & \cellcolor{red!7}45.2{\tiny($\downarrow$4.5)} & 52.6{\tiny($\downarrow$6.1)} \\
Gemma-3(4B) & 50.1{\tiny($\downarrow$4.0)} & 54.7{\tiny($\downarrow$1.9)} & 49.9{\tiny($\downarrow$2.4)} & 50.1{\tiny($\downarrow$5.3)} & \cellcolor{red!5}46.0{\tiny($\downarrow$5.0)} & 49.9{\tiny($\downarrow$5.6)} \\
Janus-Pro(7B) & \cellcolor{red!12}44.6{\tiny($\downarrow$1.8)} & 53.7{\tiny($\downarrow$0.5)} & \cellcolor{red!21}41.2{\tiny($\downarrow$3.7)} & \cellcolor{red!1}46.9{\tiny($\downarrow$1.3)} & \cellcolor{red!31}38.4{\tiny($\downarrow$2.9)} & \cellcolor{red!16}42.7{\tiny($\downarrow$0.5)} \\
Qwen2.5-VL-Instruct(3B) & \cellcolor{red!16}43.5{\tiny($\downarrow$4.2)} & 48.0{\tiny($\downarrow$3.5)} & \cellcolor{red!27}39.6{\tiny($\downarrow$6.0)} & \cellcolor{red!15}43.0{\tiny($\downarrow$4.0)} & \cellcolor{red!28}39.3{\tiny($\downarrow$7.0)} & 47.7{\tiny($\downarrow$0.7)} \\
DeepSeek-VL2 & \cellcolor{red!11}43.3{\tiny($\downarrow$2.5)} & \cellcolor{red!50}33.0{\tiny($\downarrow$1.1)} & \cellcolor{red!1}46.9{\tiny($\downarrow$2.5)} & \cellcolor{red!15}43.0{\tiny($\downarrow$1.9)} & \cellcolor{red!8}45.0{\tiny($\downarrow$3.6)} & 48.6{\tiny($\downarrow$3.4)} \\
BAGEL-MoT(7B) & \cellcolor{red!16}44.0{\tiny($\downarrow$5.6)} & 53.9{\tiny($\downarrow$2.9)} & \cellcolor{red!26}39.9{\tiny($\downarrow$8.2)} & \cellcolor{red!11}44.0{\tiny($\downarrow$5.8)} & \cellcolor{red!36}37.0{\tiny($\downarrow$5.5)} & \cellcolor{red!8}45.1{\tiny($\downarrow$5.6)} \\
ShowO2(1.5B) & \cellcolor{red!22}41.6{\tiny($\downarrow$1.8)} & 47.8{\tiny($\downarrow$1.3)} & \cellcolor{red!29}38.9{\tiny($\downarrow$1.7)} & \cellcolor{red!17}42.4{\tiny($\downarrow$2.0)} & \cellcolor{red!33}37.8{\tiny($\downarrow$2.0)} & \cellcolor{red!22}41.0{\tiny($\downarrow$1.8)} \\
\rowcolor{cyan!5}\textit{Avg.~Open-Source Models} & 55.7{\tiny($\downarrow$3.7)} & 59.8{\tiny($\downarrow$2.1)} & 54.1{\tiny($\downarrow$3.5)} & 56.4{\tiny($\downarrow$3.1)} & 51.3{\tiny($\downarrow$5.3)} & 57.1{\tiny($\downarrow$4.3)} \\

\midrule
\noalign{\vskip -\belowrulesep}
\multicolumn{7}{c}{\cellcolor{blue!15}\textbf{\textit{Medical-specialized Models}}} \\

Hulu-Med(14B) & \cellcolor{green!20}68.2{\tiny($\downarrow$3.0)} & \cellcolor{green!29}70.2{\tiny($\downarrow$2.3)} & \cellcolor{green!15}66.3{\tiny($\downarrow$2.2)} & \cellcolor{green!27}69.7{\tiny($\downarrow$1.8)} & \cellcolor{green!7}64.0{\tiny($\downarrow$4.9)} & \cellcolor{green!30}70.6{\tiny($\downarrow$4.0)} \\
UniMedVL & \cellcolor{green!13}67.1{\tiny($\downarrow$3.8)} & \cellcolor{green!49}76.1{\tiny($\downarrow$0.9)} & \cellcolor{green!12}65.3{\tiny($\downarrow$2.4)} & \cellcolor{green!15}66.1{\tiny($\downarrow$4.2)} & 60.1{\tiny($\downarrow$6.5)} & \cellcolor{green!20}67.8{\tiny($\downarrow$5.0)} \\
Hulu-Med(7B) & \cellcolor{green!9}65.4{\tiny($\downarrow$3.9)} & \cellcolor{green!28}69.9{\tiny($\downarrow$1.1)} & \cellcolor{green!5}63.5{\tiny($\downarrow$3.1)} & \cellcolor{green!13}65.8{\tiny($\downarrow$3.8)} & 61.1{\tiny($\downarrow$6.1)} & \cellcolor{green!16}66.5{\tiny($\downarrow$5.3)} \\
Lingshu(7B) & \cellcolor{green!9}65.5{\tiny($\downarrow$5.3)} & \cellcolor{green!41}73.7{\tiny($\downarrow$3.1)} & \cellcolor{green!1}62.2{\tiny($\downarrow$3.7)} & \cellcolor{green!14}65.9{\tiny($\downarrow$4.3)} & 60.0{\tiny($\downarrow$5.2)} & \cellcolor{green!13}65.6{\tiny($\downarrow$10.1)} \\
Lingshu(32B) & \cellcolor{green!6}65.1{\tiny($\downarrow$4.9)} & \cellcolor{green!42}74.0{\tiny($\downarrow$1.9)} & \cellcolor{green!1}62.1{\tiny($\downarrow$3.8)} & \cellcolor{green!11}65.1{\tiny($\downarrow$4.3)} & 57.7{\tiny($\downarrow$6.4)} & \cellcolor{green!17}66.7{\tiny($\downarrow$8.2)} \\
HealthGPT-L(14B) & 58.8{\tiny($\downarrow$2.6)} & \cellcolor{green!5}63.5{\tiny($\downarrow$2.0)} & 57.0{\tiny($\downarrow$1.6)} & 59.5{\tiny($\downarrow$1.8)} & 55.3{\tiny($\downarrow$4.1)} & 58.8{\tiny($\downarrow$3.4)} \\
MedGemma(4B) & 57.0{\tiny($\downarrow$3.3)} & \cellcolor{green!14}66.0{\tiny($\downarrow$1.8)} & 54.8{\tiny($\downarrow$1.9)} & 59.2{\tiny($\downarrow$2.5)} & 49.3{\tiny($\downarrow$5.8)} & 55.5{\tiny($\downarrow$4.5)} \\
HealthGPT-XL(32B) & 56.3{\tiny($\downarrow$3.6)} & \cellcolor{green!1}62.2{\tiny($\downarrow$2.1)} & 55.0{\tiny($\downarrow$2.6)} & 56.1{\tiny($\downarrow$2.0)} & 51.7{\tiny($\downarrow$6.4)} & 56.7{\tiny($\downarrow$4.9)} \\
HealthGPT-M(3B) & 55.5{\tiny($\downarrow$2.1)} & 58.9{\tiny($\downarrow$1.4)} & 54.7{\tiny($\downarrow$0.8)} & 55.8{\tiny($\downarrow$1.6)} & 52.3{\tiny($\downarrow$4.3)} & 56.0{\tiny($\downarrow$2.5)} \\
MedGemma(27B) & 52.2{\tiny($\downarrow$3.7)} & \cellcolor{green!1}62.2{\tiny($\downarrow$0.4)} & \cellcolor{red!0}47.4{\tiny($\downarrow$5.1)} & 52.3{\tiny($\downarrow$2.3)} & 48.8{\tiny($\downarrow$5.2)} & 50.2{\tiny($\downarrow$5.4)} \\
\rowcolor{blue!5}\textit{Avg.~Medical-Specialized} & 61.1{\tiny($\downarrow$3.6)} & 67.7{\tiny($\downarrow$1.7)} & 58.8{\tiny($\downarrow$2.7)} & 61.5{\tiny($\downarrow$2.9)} & 56.0{\tiny($\downarrow$5.5)} & 61.4{\tiny($\downarrow$5.3)} \\
\midrule
\textbf{Avg. Performance (All Models)} & \textbf{58.8}{\tiny($\downarrow$3.6)} & \textbf{63.5}{\tiny($\downarrow$2.0)} & \textbf{57.1}{\tiny($\downarrow$3.2)} & \textbf{59.2}{\tiny($\downarrow$3.0)} & \textbf{54.2}{\tiny($\downarrow$5.4)} & \textbf{59.8}{\tiny($\downarrow$4.6)} \\
\bottomrule
\end{tabular}}
\end{table}

We evaluate 40 representative MLLMs spanning three groups: 
(1) \textbf{9 \textit{Commercial MLLMs}}, including GPT-5~\cite{singh2025gpt5}, GPT-5.1~\cite{OpenAI2025GPT51}, GPT-4o-mini, GPT-4o~\cite{hurst2024gpt}, GPT-4.1-mini, GPT-4.1~\cite{OpenAI2025GPT41}, Gemini-2.5-Pro~\cite{gemini2024}, Gemini-2.5-Flash~\cite{gemini2024} and Claude-Sonnet-4.5~\cite{claude2024}; 
(2) \textbf{21 \textit{Open-source general models}}, including InternVL3-Instruct (78B, 38B)~\cite{zhu2025internvl3}, InternVL2.5-MPO (38B)~\cite{chen2025expandingperformanceboundariesopensource}, InternVL2.5 (26B)~\cite{chen2025expandingperformanceboundariesopensource}, Qwen3-VL-Instruct (235B, 30B)~\cite{qwen3vl}, Qwen2.5-VL-Instruct (72B, 32B, 3B)~\cite{bai2025qwen2}, Qwen2-VL-Instruct (72B, 7B)~\cite{wang2024qwen2}, Gemma-3 (27B, 4B)~\cite{team2025gemma}, Idefics3 (8B)~\cite{laurenccon2024building}, Mistral-Small-3.1-Instruct (24B)~\cite{MistralAI2025MistralSmall31}, ShowO2 (7B, 1.5B)~\cite{xie2025show}, BAGEL-MoT (7B)~\cite{deng2025emergingpropertiesunifiedmultimodal}, DeepSeek-VL2~\cite{wu2024deepseek}, GLM-4.5v~\cite{vteam2026glm45vglm41vthinkingversatilemultimodal}, and Janus-Pro (7B)~\cite{chen2025janusprounifiedmultimodalunderstanding}. 
(3)  \textbf{10 \textit{Medical-specialized models}}, including UniMedVL~\cite{ning2025unimedvlunifyingmedicalmultimodal}, Hulu-Med (14B, 7B)~\cite{jiang2025hulu}, Lingshu (32B, 7B)~\cite{xu2025lingshu}, HealthGPT-XL/L/M (32B, 14B, 3B)~\cite{lin2025healthgpt}, and MedGemma (27B, 4B)~\cite{sellergren2025medgemma}.

To ensure statistically significant assessments of model confidence and uncertainty, we perform $T=3$ independent inference trials for performance experiment and $T=10$ for calibration shift experiment. All models are evaluated with temperature 1.0. Detailed are provided in the supplementary material (Section B).

\subsection{Findings on Severity Degrees}
\label{ssec:exp-severity}

\textbf{Conclusion 1.} \uline{Most evaluated MLLMs exhibit severe robustness deficiency with nonlinear degradation patterns.} As shown in \cref{fig:all_models_comparison}, the majority of the MLLMs suffer significant performance drops as image quality decreases. Notably, even the best-performing model InternVL3-Instruct (78B) across all levels experiences substantial accuracy drops at a degradation severity of L2, showing that robustness deficiency remains a widespread challenge across MLLMs. 

Furthermore, the performance degradation demonstrates a concerning nonlinear acceleration pattern beyond a simple monotonic decline. While many models show some tolerance from L0 to L1 severity degrees, the performance deterioration rate increases sharply from L1 to L2. This suggests a ``cliff effect'' where model perception remains relatively stable until a certain threshold of noise is reached, after which the vision-language integration undergoes a catastrophic collapse.

\subsection{Findings on Capability Dimensions}
\label{ssec:exp-capability}

We analyze robustness across the capability dimensions. \cref{tab:capability_dimensions} provides comprehensive performance.

\textbf{Conclusion 2.} \uline{Except for a few top-tier commercial models, most MLLMs perform poorly in clinical reasoning, and existing MLLMs exhibit a certain degree of fragility across all capability dimensions.}
Across all groups, Clinical Understanding is strongest, while reasoning dimensions (Basic Science, Diagnosis, Treatment) are critically weak, with Treatment planning being the most catastrophic where multiple open-source models collapse to near-zero accuracy.

Commercial MLLMs substantially outperform open-source models in clinical reasoning, whereas leading open-source models remain competitive in Medical Perception, even surpassing commercial models in some dimensions.
Medical-specialized models show no consistent advantage over general models.
Regarding robustness under degradation, Treatment planning is the most resilient capability with the smallest average performance drop, while Anatomical Recognition exhibits the poorest robustness with the largest degradation magnitude; the remaining dimensions show broadly comparable levels of degradation sensitivity.

\subsection{Findings on Degradation-Type Sensitivity}
\label{ssec:exp-degradation}

We analyze model sensitivity to different degradation types. Comprehensive degradation heatmaps showing performance drops from baseline (L0) to gradual degradations (L1 and L2) across different models and degradation types are presented in \cref{tab:degradation_categories}. 

\textbf{Conclusion 3.} \uline{MLLMs are strikingly vulnerable to physics-based artifacts and motion interference compared to intensity-based degradations.}
The five degradation categories produce markedly different impacts on model robustness. Degradations involving ``intensity jitter'' demonstrate the highest resilience with the smallest performance drops, while ``noise'' and ``resolution \& blur'' show moderate impact. In sharp contrast, ``artifact'' degradations, which encompass medical physics-specific corruptions such as MRI undersampling artifacts and sparse-view CT artifacts, cause larger drops. Motion-induced degradations such as object rotation, also show substantial impact, ranking second in severity after artifacts. This further indicates that medical artifacts and motion interference are substantially more destructive in corrupting the image semantics than other types of degradations.

This vulnerability hierarchy reveals that current MLLMs lack sufficient understanding of medical imaging-specific degradation mechanisms that are absent from natural image pretraining distributions. 
 Besides, modality-specific sensitivity patterns are further analyzed in the supplementary material (Section C).

\subsection{Findings on Overconfidence Under Degradations}
\label{ssec:exp-overconfidence}

\begin{figure*}[t]
  \centering
  \includegraphics[width=\columnwidth]{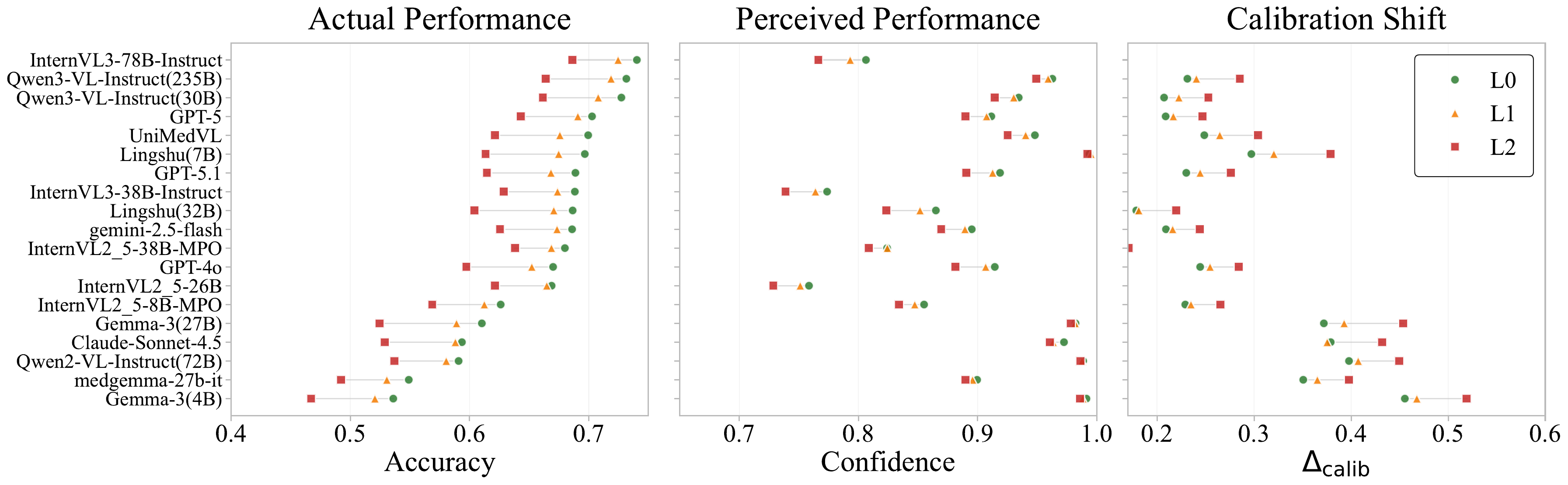}
  \caption{\textbf{Comprehensive analysis of model calibration shift across severity degrees.} These panels quantify the widening gap between actual accuracy and model certainty under increasing image degradation. As degradation scales from L0 to L2, the stability of perceived performance against collapsing accuracy reveals a fundamental failure in metacognitive awareness across MLLMs.}
  \label{fig:overconfidence_three_panels}
  \end{figure*}

Furthermore, to investigate the metacognitive ability of MLLMs, we analyze the relationship between actual performance and model confidence under degradation. Following the Calibration Shift proposed in Section~\ref{ssec:bench-metrics}, we measure the gap between perceived confidence and actual performance.

\textbf{Conclusion 4.} \uline{All models exhibit severe overconfidence under degradation, demonstrating the AI Dunning-Kruger Effect.} As shown in \cref{fig:overconfidence_three_panels}, models maintain inappropriately high perceived confidence despite substantial accuracy degradation. Across all models and quality degrees, calibration shift remains consistently positive and increases systematically from L0 to L2. 

Our analysis reveals this phenomenon manifests in two dimensions. \textit{\uline{Intra-Model DKE}} characterizes metacognitive failure within individual models: as degradation severity increases from L0 to L2, actual performance drops substantially while calibration shift consistently increases, indicating models grow more overconfident precisely when their capabilities deteriorate. This pattern holds universally across all evaluated architectures. \textit{\uline{Inter-Model DKE}} describes the inverse relationship across different models: at severe degradation degrees, lower-performing models overall exhibit disproportionately higher calibration shift compared to higher-capability models.

\begin{figure}[!t]
  \centering
  \begin{subfigure}[b]{0.50\linewidth}
      \centering
      \includegraphics[width=\linewidth]{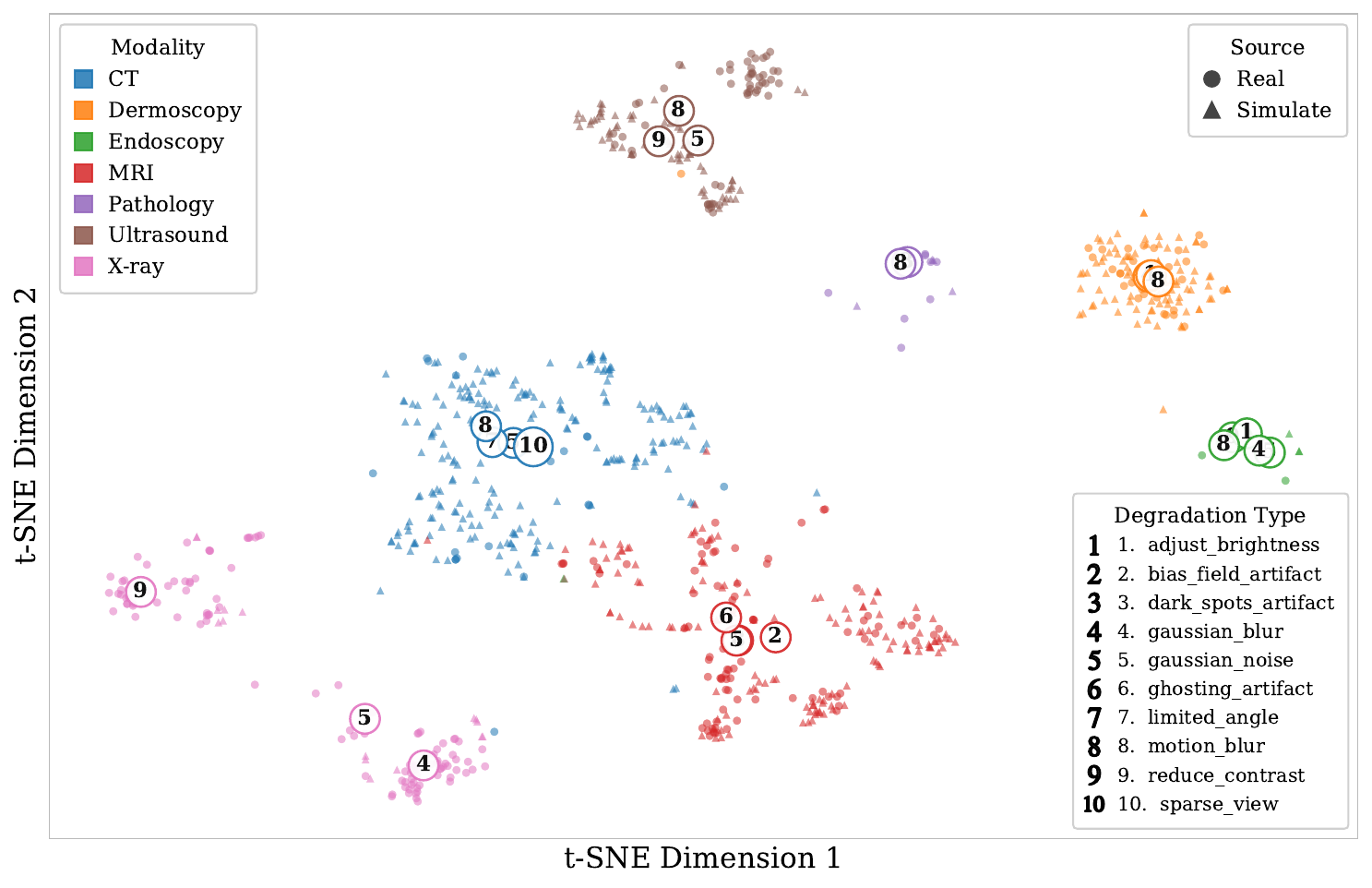}
      \caption{}
      \label{fig:tsne_degtype}
  \end{subfigure}
  \hfill
  \begin{subfigure}[b]{0.48\linewidth}
      \centering
      \includegraphics[width=\linewidth]{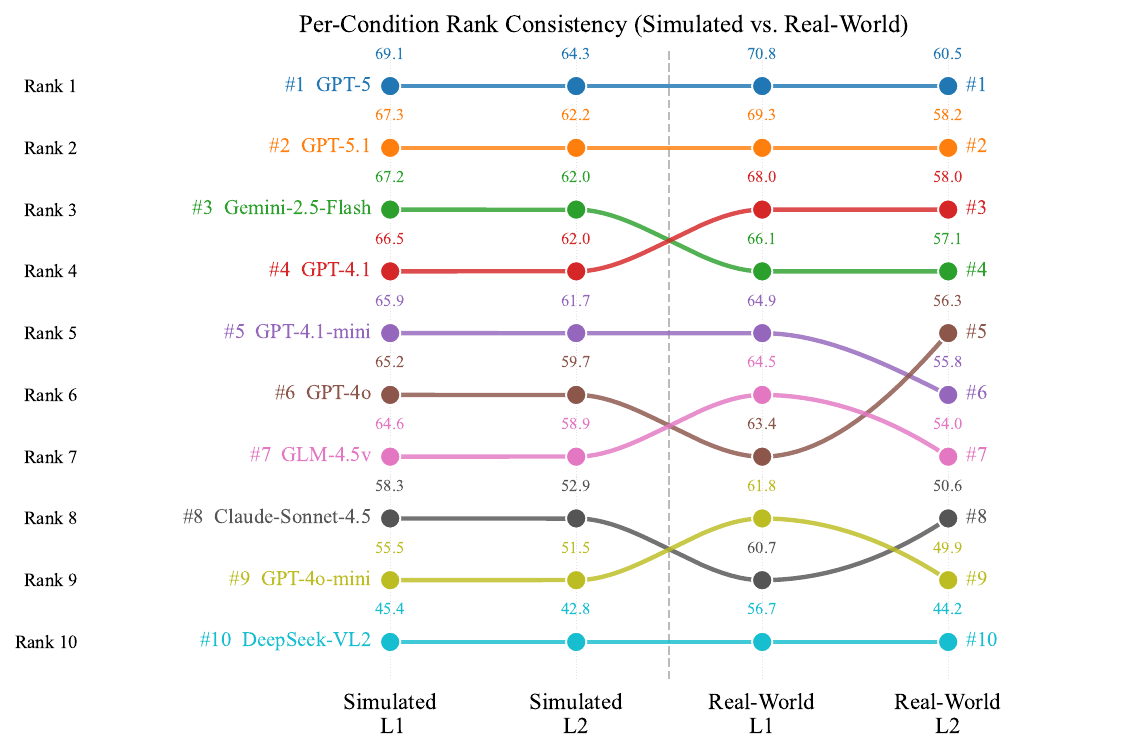}
      \caption{}
      \label{fig:sim_real_bump}
  \end{subfigure}
  \caption{\textbf{Simulated vs.\ Real-World Evaluation Consistency.}
  \textit{Left:} t-SNE projection of BiomedCLIP features from simulated and real degraded images.
  Real and simulated images co-locate within each modality cluster, confirming distribution alignment.
  \textit{Right:} Bump chart tracing per-model ranks across Simulated-Mild, Simulated-Severe, Real-World-Mild, and Real-World-Severe conditions.}
  \label{fig:sim_real_consistency}
\end{figure}

\subsection{Distribution Alignment Validation}
\label{app:distribution_alignment}

A key concern for any benchmark that includes simulated images is whether the simulated degradations are representative of those encountered in real-world clinical practice.
We address this question through two complementary analyses: (1) a feature-space distribution comparison using t-SNE analysis~\cite{van2008visualizing}, and (2) a cross-evaluation rank consistency study.

\textbf{Feature Distribution Analysis.}
We extract feature vectors from 2{,}000 simulated and 2{,}000 real-world degraded images using BiomedCLIP~\cite{zhang2023biomedclip}, a vision encoder pretrained on large-scale biomedical image-text pairs. We then project these feature vectors into a two-dimensional space using t-SNE.  
As shown in \cref{fig:sim_real_consistency} (a), \textit{simulated and real degraded images exhibit highly overlapping feature distributions within each modality cluster}, suggesting that the degradations simulated by our pipeline capture characteristics similar to those observed in real-world data.

\textbf{Rank Consistency Analysis.}
We further evaluate whether model rankings on simulated images faithfully predict rankings on 2{,}000 real-world degraded clinical QA.
As shown in \cref{fig:sim_real_consistency} (b), \textit{per-model rank trajectories remain highly consistent across Simulated-Mild, Simulated-Severe, Real-World-Mild, and Real-World-Severe conditions.} 
Only a few adjacent-rank swaps are observed (\eg, GPT-4.1 and Gemini-2.5-Flash), while the global ordering remains stable, with GPT-5 and GPT-5.1 dominating and DeepSeek-VL2 consistently at the bottom.
Together, these results confirm that \benchname evaluation scores obtained on simulated degradations are reliable proxies for real-world clinical performance rankings.

\section{Discussion}

This paper presents \textbf{\benchname}, a large-scale and Multidimensional benchmark for evaluating medical MLLMs under clinically realistic image quality degradations. Our evaluation of 40 models reveals that robustness deficiency is both universal and nonlinear: most models tolerate mild corruptions (L0 to L1) reasonably well but collapse sharply at severity degrees (L1 to L2), suggesting a fundamental fragility in vision-language integration beyond a certain degradation threshold. 
\uline{Capability-level analysis} (Section~\ref{ssec:exp-capability}) reveals that most models already perform poorly in clinical reasoning tasks at baseline, and this weakness persists under degradation. Notably, Anatomical Recognition exhibits the poorest degradation robustness despite being a perception-oriented task, while Treatment planning shows the greatest resilience, suggesting that degradation sensitivity is shaped more by the granularity of visual detail required than by whether a task involves reasoning or perception. 
\uline{At the degradation level} (Section~\ref{ssec:exp-degradation}), physics-based artifacts and motion interference cause substantially larger performance drops than intensity or resolution perturbations, consistent with the absence of such domain-specific corruptions from natural image pretraining distributions. 

Perhaps the most clinically alarming finding is the  \uline{AI Dunning-Kruger Effect} (Section~\ref{ssec:exp-overconfidence}): as degradation severity increases, model accuracy collapses while perceived confidence remains high. This metacognitive failure holds universally across all 40 architectures and manifests in both intra-model and inter-model forms, posing direct patient safety risks by preventing appropriate human oversight from being triggered. Developing models that are not only accurate but also well-calibrated under degradation is therefore a critical open problem that we hope \benchname will help drive.

A potential concern for simulation-based benchmarks is ecological validity. Our distribution alignment analysis (Section~\ref{app:distribution_alignment}) confirms that simulated and real degraded images are nearly indistinguishable in BiomedCLIP feature space, and that model rankings on simulated data faithfully predict rankings on real clinical images, \uline{validating \benchname as a reliable proxy for real-world robustness assessment}. \textit{We believe \benchname provides essential infrastructure for developing medical MLLMs that are not only accurate on clean data, but reliable under the imperfect conditions of real clinical practice.}

% \clearpage  % TODO FINAL: This \clearpage needs to be removed from both review and camera-ready versions.

\bibliographystyle{splncs04}
\bibliography{main}

\appendix
\clearpage

% ---------------------------------------------------------------
% Supplementary Material (merged from appendix.tex)
% ---------------------------------------------------------------

\renewcommand{\contentsname}{Supplementary Material of \benchname}
% \setcounter{tocdepth}{2}
% {\small\tableofcontents}
% \vspace{1em}

\section{Dataset Details}

\subsection{Expert Annotation Interface}
\label{app:annotation_interface}

\begin{figure}[!ht]
    \centering
    \includegraphics[width=\linewidth]{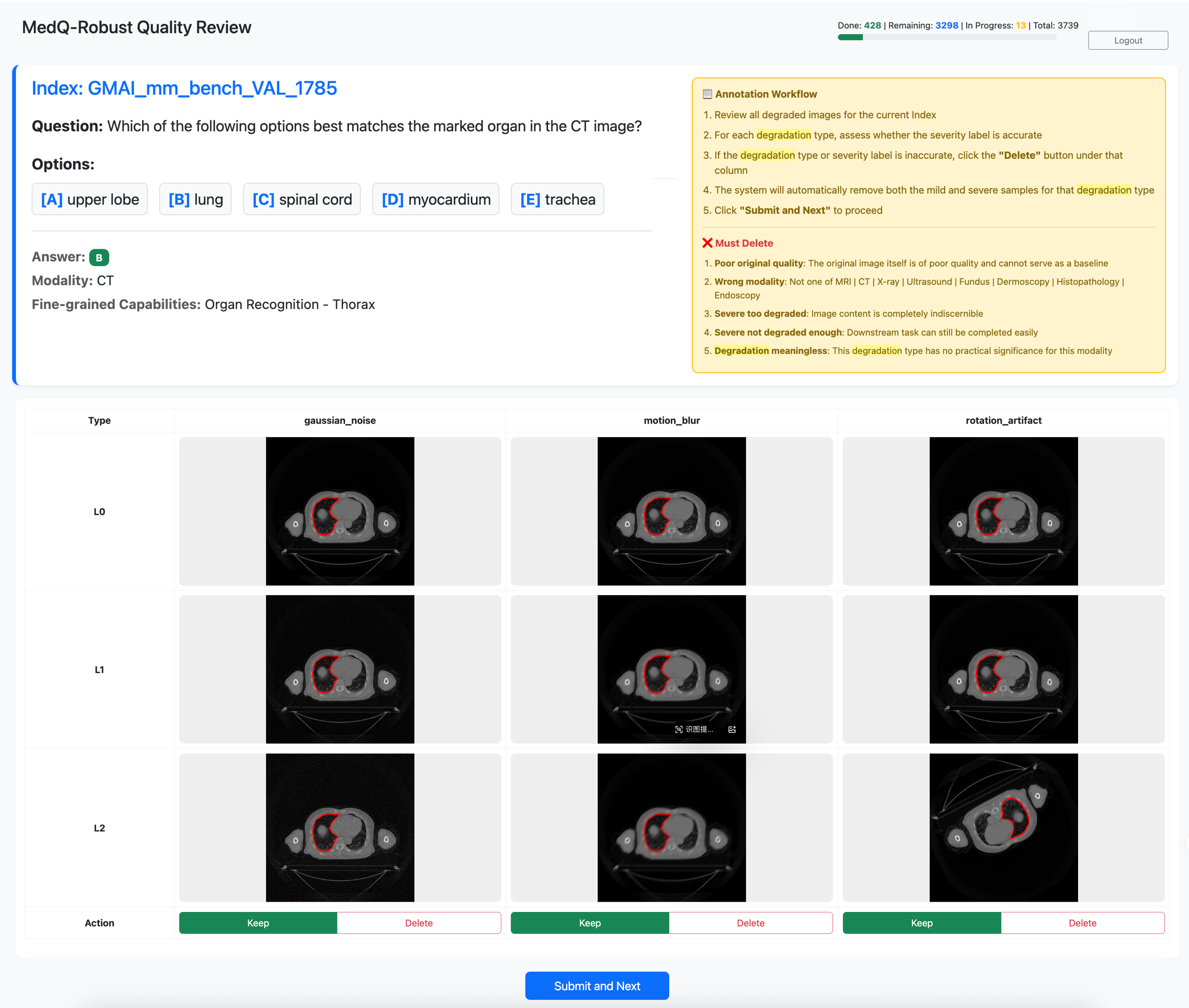}
    \caption{\textbf{Expert annotation interface.}}
    \label{fig:annotation_interface}
\end{figure}

To ensure the clinical validity of \benchname, three board-certified radiologists independently reviewed all generated degraded image--question pairs through a structured annotation interface, as illustrated in Figure~\ref{fig:annotation_interface}.
Annotators were presented with samples in randomized order to minimize ordering bias.
For each sample, annotators made one of two decisions:
\begin{itemize}
  \item \underline{Retain}: If the degradation type, severity assignment, and image--question relevance are all deemed appropriate, the annotator proceeds by clicking ``Next''.
  \item \underline{Discard}: If any of the following issues are identified (incorrect severity assignment, mislabeled degradation type, or lack of visual relevance between the question and the image), the sample is removed from the dataset.
\end{itemize}

Mandatory discard criteria are defined as follows:
\begin{enumerate}
  \item \underline{Poor baseline quality}: The original (L0) image is itself of insufficient diagnostic quality to serve as a clean reference.
  \item \underline{Modality mismatch}: The image does not belong to any of the seven target modalities: MRI, CT, X-ray, Ultrasound, Dermoscopy, Histopathology, or Endoscopy.
  \item \underline{Severe over-degradation}: The L2 image is so severely corrupted that no discernible diagnostic content remains.
  \item \underline{Insufficient degradation at L2}: The L2 image remains visually indistinguishable from the clean reference, providing no meaningful visual challenge.
  \item \underline{Clinically irrelevant degradation}: The assigned degradation type has no practical occurrence or clinical significance for the given imaging modality.
\end{enumerate}

Samples failing any of the above criteria were excluded from \benchname.
In total, approximately 8.3\% of generated pairs were discarded through this process.

\subsection{Complete Capability Dimension Hierarchy}
\label{app:capability_taxonomy}

Table~\ref{tab:fine_grained_capability} expands the three-level capability hierarchy summarized in the main paper into its full fine-grained structure, listing all 30 leaf-level capabilities with their mid-level groupings and QA counts.

\begin{table*}[!ht]
    \centering
    \footnotesize
    \caption{\textbf{Capability hierarchy of \benchname.} Row shading indicates high-level capability: \colorbox{SkyBlue!25}{blue} rows belong to \textit{Medical Perception} and \colorbox{Violet!15}{purple} rows belong to \textit{Clinical Reasoning}.}
    \resizebox{\textwidth}{!}{
    \renewcommand{\arraystretch}{0.9}  % 减小行高
    \begin{tabular}{lcccc}
    \toprule
    \textbf{Fine-grained Capability} & \textbf{Mid-level Capability} & \textbf{Count} & \textbf{Ratio in Mid-level} & \textbf{Ratio in Total} \\
    \midrule
    \rowcolor{SkyBlue!25} Disease Diagnosis & Clinical Understanding & 12,908 & 97.74\% & 62.70\% \\
    \rowcolor{SkyBlue!25} Severity Grading & Clinical Understanding & 138 & 1.04\% & 0.67\% \\
    \rowcolor{SkyBlue!25} Basic Science Understanding & Clinical Understanding & 78 & 0.59\% & 0.38\% \\
    \rowcolor{SkyBlue!25} Lesion Grading & Clinical Understanding & 68 & 0.51\% & 0.33\% \\
    \rowcolor{SkyBlue!25} Attribute Recognition & Clinical Understanding & 10 & 0.08\% & 0.05\% \\
    \rowcolor{SkyBlue!25} Treatment-Understanding & Clinical Understanding & 4 & 0.03\% & 0.02\% \\
    \midrule
    \rowcolor{SkyBlue!25} Counting & Imaging Perception & 390 & 73.58\% & 1.89\% \\
    \rowcolor{SkyBlue!25} Surgical Workflow Recognition & Imaging Perception & 96 & 18.11\% & 0.47\% \\
    \rowcolor{SkyBlue!25} Image Quality Grading & Imaging Perception & 28 & 5.28\% & 0.14\% \\
    \rowcolor{SkyBlue!25} Surgical Instrument Recognition & Imaging Perception & 16 & 3.02\% & 0.08\% \\
    \midrule
    \rowcolor{SkyBlue!25} Skeletal System & Anatomical Recognition & 2,562 & 41.22\% & 12.45\% \\
    \rowcolor{SkyBlue!25} Cardiovascular & Anatomical Recognition & 1,392 & 22.39\% & 6.76\% \\
    \rowcolor{SkyBlue!25} Abdominal Structures & Anatomical Recognition & 880 & 14.16\% & 4.27\% \\
    \rowcolor{SkyBlue!25} Respiratory System & Anatomical Recognition & 726 & 11.68\% & 3.53\% \\
    \rowcolor{SkyBlue!25} Gastrointestinal Tract & Anatomical Recognition & 475 & 7.64\% & 2.31\% \\
    \rowcolor{SkyBlue!25} Neurovascular & Anatomical Recognition & 463 & 7.45\% & 2.25\% \\
    \rowcolor{SkyBlue!25} General Structures & Anatomical Recognition & 328 & 5.28\% & 1.59\% \\
    \rowcolor{SkyBlue!25} Muscle Recognition & Anatomical Recognition & 231 & 3.72\% & 1.12\% \\
    \rowcolor{SkyBlue!25} Cell and Microbes & Anatomical Recognition & 158 & 2.54\% & 0.77\% \\
    \midrule
    \rowcolor{Violet!15} General Science Reasoning & Basic Science Reasoning & 74 & 71.15\% & 0.36\% \\
    \rowcolor{Violet!15} Pathophysiological Reasoning & Basic Science Reasoning & 26 & 25.00\% & 0.13\% \\
    \rowcolor{Violet!15} Clinical Correlation & Basic Science Reasoning & 4 & 3.85\% & 0.02\% \\
    \midrule
    \rowcolor{Violet!15} General Diagnosis Reasoning & Diagnosis Reasoning & 336 & 87.50\% & 1.63\% \\
    \rowcolor{Violet!15} Laboratory/Imaging Interpretation & Diagnosis Reasoning & 38 & 9.90\% & 0.18\% \\
    \rowcolor{Violet!15} Clinical Presentation Analysis & Diagnosis Reasoning & 6 & 1.56\% & 0.03\% \\
    \rowcolor{Violet!15} Disease Mechanism Recognition & Diagnosis Reasoning & 4 & 1.04\% & 0.02\% \\
    \midrule
    \rowcolor{Violet!15} Treatment Reasoning & Treatment Reasoning & 127 & 85.81\% & 0.62\% \\
    \rowcolor{Violet!15} Pharmacological Treatment & Treatment Reasoning & 12 & 8.11\% & 0.06\% \\
    \rowcolor{Violet!15} Initial/Emergency Management & Treatment Reasoning & 7 & 4.73\% & 0.03\% \\
    \rowcolor{Violet!15} Surgical Treatment Selection & Treatment Reasoning & 2 & 1.35\% & 0.01\% \\
    \bottomrule
    \end{tabular}
    }
    \label{tab:fine_grained_capability}
    \vspace{-5pt}
\end{table*}

\subsection{Complete Degradation Taxonomy}
\label{app:degradation_taxonomy}

Table~\ref{tab:degradation_stats} reports the complete statistics for all 18 degradation types across the five top-level categories, including sample counts and proportions at both the category and dataset level.

\begin{table*}[!ht]
    \centering
    \footnotesize
    \caption{\textbf{Degradation type statistics.}}
    \resizebox{\textwidth}{!}{
    \begin{tabular}{llcccl}
    \toprule
    \textbf{Name} & \textbf{Parent Category} & \textbf{Count} & \textbf{Ratio in Parent} & \textbf{Ratio in Total} & \textbf{Modality} \\
    \midrule
    \rowcolor{gray!15} \textbf{Artifacts} & - & \textbf{5,322} & - & \textbf{25.85\%} & \textbf{All} \\
    limited\_angle & Artifacts & 1,440 & 27.06\% & 6.99\% & CT \\
    sparse\_view & Artifacts & 1,440 & 27.06\% & 6.99\% & CT \\
    bias\_field\_artifact & Artifacts & 936 & 17.59\% & 4.55\% & MRI \\
    undersampling\_artifact & Artifacts & 672 & 12.63\% & 3.26\% & MRI \\
    ghosting\_artifact & Artifacts & 352 & 6.61\% & 1.71\% & MRI \\
    blood\_cell\_artifact & Artifacts & 304 & 5.71\% & 1.48\% & Pathology \\
    dark\_spots\_artifact & Artifacts & 178 & 3.34\% & 0.86\% & Pathology \\
    \midrule
    \rowcolor{gray!15} \textbf{Motion Interference} & - & \textbf{2,992} & - & \textbf{14.53\%} & \textbf{All} \\
    object\_rotation & Motion & 2,718 & 90.84\% & 13.20\% & All \\
    object\_movement & Motion & 274 & 9.16\% & 1.33\% & All \\
    \midrule
    \rowcolor{gray!15} \textbf{Intensity Jitter} & - & \textbf{1,846} & - & \textbf{8.97\%} & \textbf{All} \\
    adjust\_brightness & Intensity & 916 & 49.62\% & 4.45\% & All \\
    exposure & Intensity & 518 & 28.06\% & 2.52\% & All \\
    reduce\_contrast & Intensity & 412 & 22.32\% & 2.00\% & All \\
    \midrule
    \rowcolor{gray!15} \textbf{Noise} & - & \textbf{3,796} & - & \textbf{18.44\%} & \textbf{All} \\
    gaussian\_noise & Noise & 2,680 & 70.60\% & 13.02\% & All \\
    low\_dose & Noise & 1,116 & 29.40\% & 5.42\% & CT \\
    \midrule
    \rowcolor{gray!15} \textbf{Resolution \& Blur} & - & \textbf{6,632} & - & \textbf{32.21\%} & \textbf{All} \\
    low\_resolution & Resolution and Blur & 3,046 & 45.93\% & 14.80\% & All \\
    motion\_blur & Resolution and Blur & 2,820 & 42.52\% & 13.70\% & All \\
    gaussian\_blur & Resolution and Blur & 470 & 7.09\% & 2.28\% & All \\
    bubble & Resolution and Blur & 296 & 4.46\% & 1.44\% & Pathology \\
    \bottomrule
    \end{tabular}
    }
    \label{tab:degradation_stats}
    \vspace{-5pt}
\end{table*}

\subsection{Degradation Type Definitions}
\label{app:degradation_definitions}

Below we provide concise definitions for each of the 18 degradation variants in \benchname, organized by the five top-level categories. Types are further annotated as \textit{general} (applicable across all modalities) or \textit{modality-specific}.

\noindent\textbf{Artifacts (7 types, modality-specific).}
\underline{Limited-angle reconstruction} (CT): simulates under-complete angular sampling during CT acquisition by discarding a fraction of projection angles, producing streak and shadow artifacts.
\underline{Sparse-view reconstruction} (CT): simulates dose-reduced CT by sub-sampling raw sinogram measurements at regular angular intervals, resulting in ring artifacts and resolution loss.
\underline{Bias field artifact} (MRI): applies a smooth, spatially varying intensity inhomogeneity field, mimicking $B_1$ field non-uniformity in clinical MRI scanners.
\underline{Undersampling artifact} (MRI): simulates accelerated $k$-space acquisition by retaining only a fraction of frequency-domain measurements, causing aliasing and ghosting.
\underline{Ghosting artifact} (MRI): replicates periodic motion-induced Gibbs ringing along the phase-encode direction, a common artefact from respiratory or cardiac motion.
\underline{Blood cell artifact} (Histopathology): synthetically overlays erythrocyte-like circular occlusions onto tissue sections, mimicking contamination from hemorrhage during slide preparation.
\underline{Dark spot artifact} (Histopathology): introduces irregularly shaped dark regions that mimic air bubbles, folded tissue, or staining precipitates common in whole-slide imaging.

\noindent\textbf{Motion Interference (2 types, general).}
\underline{Object rotation}: applies an in-plane affine rotation of random magnitude to simulate patient repositioning or gantry misalignment artifacts.
\underline{Object movement}: applies a random in-plane translation, simulating patient displacement between scan sequences.

\noindent\textbf{Intensity Jitter (3 types, general).}
\underline{Brightness adjustment}: uniformly shifts pixel intensity values to simulate over- or under-exposure in acquisition.
\uline{Exposure variation}: applies a non-linear gamma correction, mimicking variations in sensor gain or automatic exposure control.
\underline{Contrast reduction}: compresses the dynamic range of intensity values, simulating low-contrast imaging conditions from suboptimal window/level settings.

\noindent\textbf{Noise (2 types).}
\underline{Gaussian noise} (general): adds zero-mean isotropic Gaussian noise, modeling thermal noise from detector electronics.
\underline{Low-dose noise} (CT, modality-specific): simulates photon-count noise from dose-reduced CT acquisition using a Poisson noise model applied to the sinogram domain.

\noindent\textbf{Resolution \& Blur (4 types).}
\underline{Low resolution} (general): downsamples the image by a scale factor and upsamples back to original size, simulating reduced spatial resolution from hardware or acquisition protocol constraints.
\underline{Motion blur} (general): applies a directional convolution kernel, modeling camera or patient motion integrated over the exposure window.
\underline{Gaussian blur} (general): applies an isotropic Gaussian low-pass filter, simulating optical defocus or post-processing smoothing.
\underline{Bubble} (Histopathology): overlays semi-transparent circular regions mimicking air bubbles trapped beneath the coverslip during slide mounting.

\section{Evaluation}

This section details the inference protocol used to evaluate all models.

\noindent\textbf{Prompt template.}
All 40 models are evaluated under a unified multiple-choice prompting protocol. Given a degraded image $\mathcal{I}_c$ and a question with $K$ answer options, the prompt is formatted as:

\begin{Verbatim}[frame=single, fontsize=\small, framesep=3mm]
You are a medical AI assistant. Please answer the following
question based on the provided medical image. {question}

{option_text}

Constraint: Output ONLY the single letter (A, B, C, or D,
E, etc) corresponding to the correct answer. No explanation,
no punctuation.

Answer:
\end{Verbatim}

\noindent\textbf{Inference settings.}
All open-source models are deployed on NVIDIA A100 GPUs.
Model outputs are matched to answer options using a priority rule: (1) direct single-letter match; (2) first capital letter found in response; (3) if no letter is identified, the sample is treated as incorrect.

\clearpage
\section{Results}

\subsection{Overall Model Performance}\label{detail_res}

Table~\ref{tab:main_results} reports the full per-model accuracy at each severity degree (L0, L1, L2) along with the absolute drops relative to the clean baseline, covering all 40 models evaluated in \benchname.

\begin{table*}[!ht]
    \centering
    \footnotesize
    % \tiny
    \setlength{\tabcolsep}{16pt}
    \renewcommand{\arraystretch}{0.9}
    \caption{\textbf{Overall accuracy across severity degrees.} L0/L1/L2 accuracies and drops (L1-L0, L2-L0).}
    \resizebox{1.0\textwidth}{!}{
    \begin{tabular}{lccccc}
    \toprule
    \textbf{Model} & \textbf{L0$\uparrow$} & \textbf{L1$\uparrow$} & \textbf{L2$\uparrow$} & \textbf{L1-L0$\downarrow$} & \textbf{L2-L0$\downarrow$}  \\
    \midrule
    \multicolumn{6}{c}{\cellcolor{orange!15}\textbf{\textit{Commercial MLLMs}}} \\
    Gemini-2.5-Pro & 71.39 & 69.93 & 65.20 & -1.46 & -6.19 \\
    GPT-5 & 70.27 & 69.09 & 64.30 & -1.18 & -5.97 \\
    GPT-5.1 & 68.56 & 67.34 & 62.16 & -1.22 & -6.40 \\
    Gemini-2.5-Flash & 67.39 & 67.18 & 62.00 & -0.21 & -5.39 \\
    GPT-4o & 66.98 & 65.22 & 59.73 & -1.76 & -7.25 \\
    GPT-4.1 & 66.00 & 66.50 & 62.00 & 0.50 & -4.00 \\
    GPT-4.1-mini & 65.21 & 65.87 & 61.67 & 0.66 & -3.54 \\
    Claude-Sonnet-4.5 & 59.64 & 58.28 & 52.86 & -1.36 & -6.78 \\
    GPT-4o-mini & 57.32 & 55.49 & 51.53 & -1.83 & -5.79 \\
    \midrule
    \multicolumn{6}{c}{\cellcolor{cyan!15}\textbf{\textit{Open-Source General Models}}} \\
    InternVL3-Instruct(78B) & 74.04 & 72.46 & 68.63 & -1.58 & -5.41 \\
    Qwen3-VL-Instruct(235B) & 73.15 & 71.87 & 66.40 & -1.28 & -6.75 \\
    Qwen3-VL-Instruct(30B) & 72.74 & 70.79 & 66.14 & -1.95 & -6.60 \\
    InternVL3-Instruct(38B) & 68.83 & 67.38 & 62.87 & -1.45 & -5.96 \\
    InternVL2.5-MPO(38B) & 68.00 & 66.87 & 63.81 & -1.13 & -4.19 \\
    InternVL2.5(26B) & 66.88 & 66.48 & 62.12 & -0.40 & -4.76 \\
    GLM-4.5v & 65.77 & 64.57 & 58.94 & -1.20 & -6.83 \\
    Gemma-3(27B) & 61.03 & 58.91 & 52.45 & -2.12 & -8.58 \\
    Qwen2.5-VL-Instruct(72B) & 59.68 & 58.28 & 52.92 & -1.40 & -6.76 \\
    Qwen2-VL-Instruct(72B) & 58.52 & 58.26 & 53.61 & -0.26 & -4.91 \\
    Qwen2.5-VL-Instruct(32B) & 57.22 & 55.67 & 50.69 & -1.55 & -6.53 \\
    Idefics3(8B) & 56.71 & 55.85 & 51.82 & -0.86 & -4.89 \\
    Qwen2-VL-Instruct(7B) & 55.69 & 54.77 & 51.14 & -0.92 & -4.55 \\
    ShowO2(7B) & 55.60 & 53.59 & 47.72 & -2.01 & -7.88 \\
    Gemma-3(4B) & 53.60 & 52.06 & 46.70 & -1.54 & -6.90 \\
    Mistral-Small-3.1-Instruct(24B) & 53.55 & 53.43 & 48.61 & -0.12 & -4.94 \\
    BAGEL-MoT(7B) & 48.30 & 45.17 & 39.92 & -3.13 & -8.38 \\
    Qwen2.5-VL-Instruct(3B) & 46.86 & 44.72 & 40.43 & -2.14 & -6.43 \\
    DeepSeek-VL2 & 46.77 & 45.37 & 42.80 & -1.40 & -3.97 \\
    Janus-Pro(7B) & 45.29 & 45.23 & 42.17 & -0.06 & -3.12 \\
    ShowO2(1.5B) & 43.15 & 42.47 & 39.35 & -0.68 & -3.80 \\
    \midrule
    \multicolumn{6}{c}{\cellcolor{blue!15}\textbf{\textit{Medical-Specialized Models}}} \\
    UniMedVL & 70.18 & 68.48 & 62.69 & -1.70 & -7.49 \\
    Hulu-Med(14B) & 70.04 & 70.20 & 65.40 & 0.16 & -4.64 \\
    Lingshu(7B) & 69.67 & 67.48 & 61.34 & -2.19 & -8.33 \\
    Lingshu(32B) & 68.65 & 67.07 & 60.40 & -1.58 & -8.25 \\
    Hulu-Med(7B) & 68.09 & 67.07 & 62.20 & -1.02 & -5.89 \\
    HealthGPT-L(14B) & 60.29 & 59.71 & 56.75 & -0.58 & -3.54 \\
    HealthGPT-XL(32B) & 58.99 & 57.16 & 53.69 & -1.83 & -5.30 \\
    MedGemma(4B) & 58.48 & 58.12 & 53.77 & -0.36 & -4.71 \\
    HealthGPT-M(3B) & 57.32 & 56.61 & 53.45 & -0.71 & -3.87 \\
    MedGemma(27B) & 54.90 & 53.06 & 49.22 & -1.84 & -5.68 \\
    \bottomrule
    \end{tabular}
    }
    \label{tab:main_results}
\end{table*}

\subsection{Modality-Specific Performance Analysis}
\label{app:modality_performance}

\begin{figure}[!ht]
    \centering
    \includegraphics[width=\columnwidth]{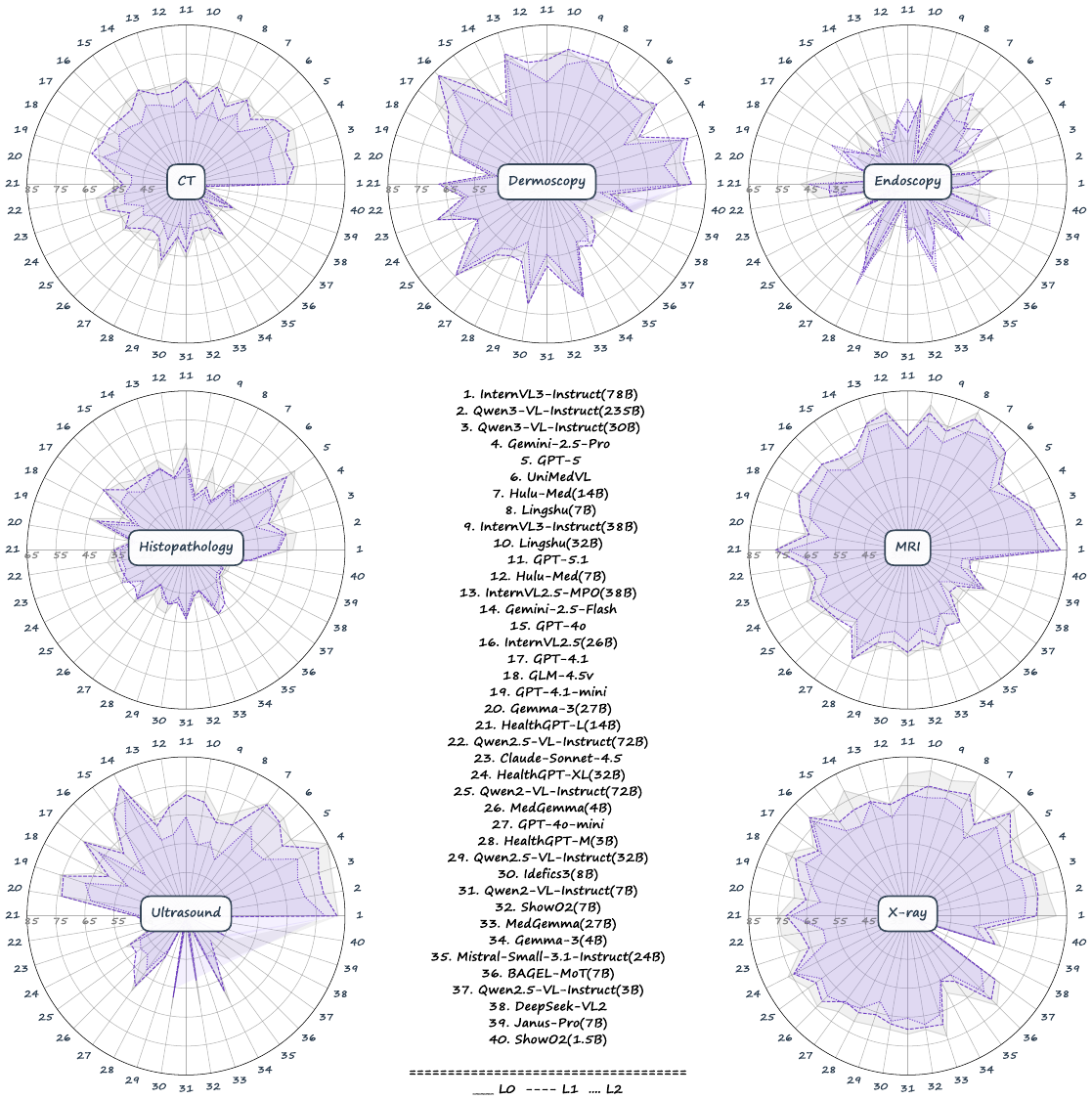}
    \caption{\textbf{Performance comparison across seven medical imaging modalities under three severity degrees.} Each radar chart shows model performance for a specific modality.}
    \label{fig:radar_combined_modalities}
    \vspace{-3em}
\end{figure}

Figure~\ref{fig:radar_combined_modalities} visualizes per-modality performance profiles for all models across all three severity degrees.

\subsection{Performance Heatmap}
\label{app:heatmap}

\begin{figure}[!ht]
    \centering
    \includegraphics[width=\linewidth]{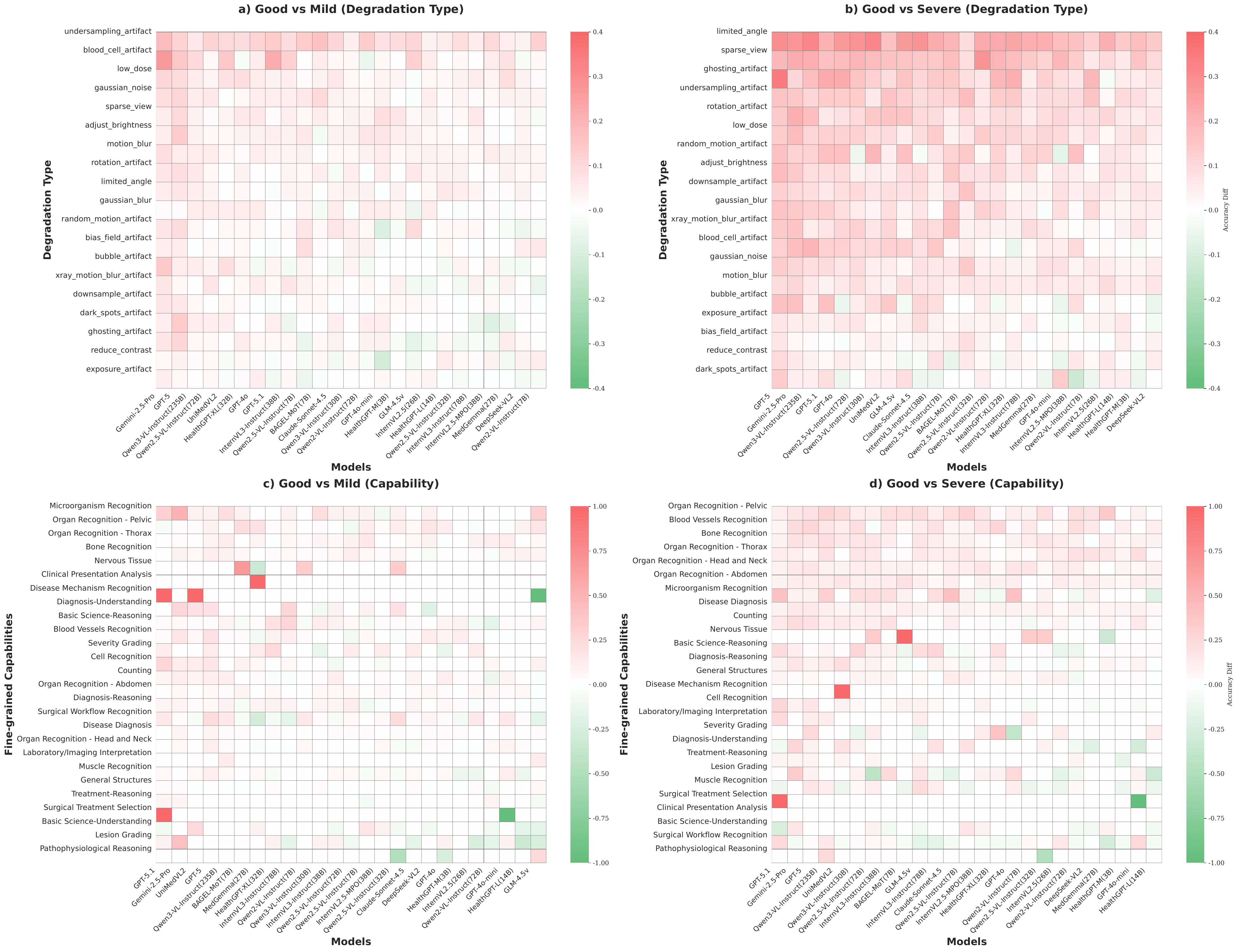}
    \caption{\textbf{Accuracy difference heatmaps (L0 $-$ Degraded).}
    \textcolor[HTML]{f8696b}{Red}: degradation hurts accuracy; \textcolor[HTML]{63be7b}{green}: no effect or slight gain; white: no change.
    \textbf{(a)} L0 vs.\ L1 by degradation type.
    \textbf{(b)} L0 vs.\ L2 by degradation type.
    \textbf{(c)} L0 vs.\ L1 by fine-grained capability.
    \textbf{(d)} L0 vs.\ L2 by fine-grained capability.}
    \label{fig:heatmap_combined}
    % \vspace{-1.5em}
\end{figure}

Figure~\ref{fig:heatmap_combined} shows accuracy differences (L0 $-$ degraded) across models, degradation types, and fine-grained capabilities at mild (L1) and severe (L2) levels.

% \clearpage
\subsection{Visual Examples of Image Degradations}
\label{app:degradation_examples}

\begin{figure}[!ht]
    \centering
    \includegraphics[width=\linewidth]{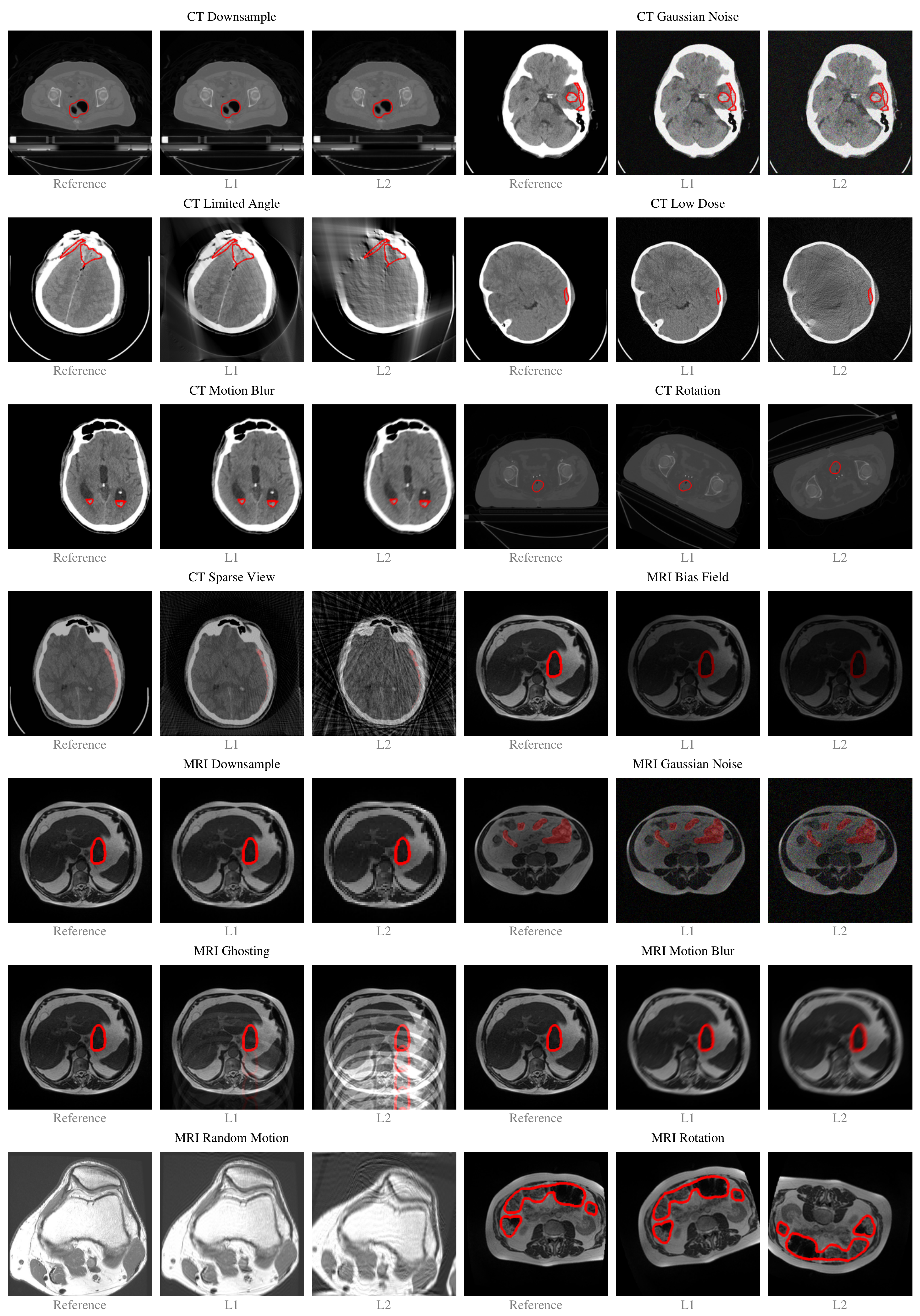}
    \caption{\textbf{Degradation examples (Part 1/3): CT and MRI modalities.}}
    \label{fig:degradation_examples_p1}
\end{figure}

\begin{figure}[!ht]
    \centering
    \includegraphics[width=\linewidth]{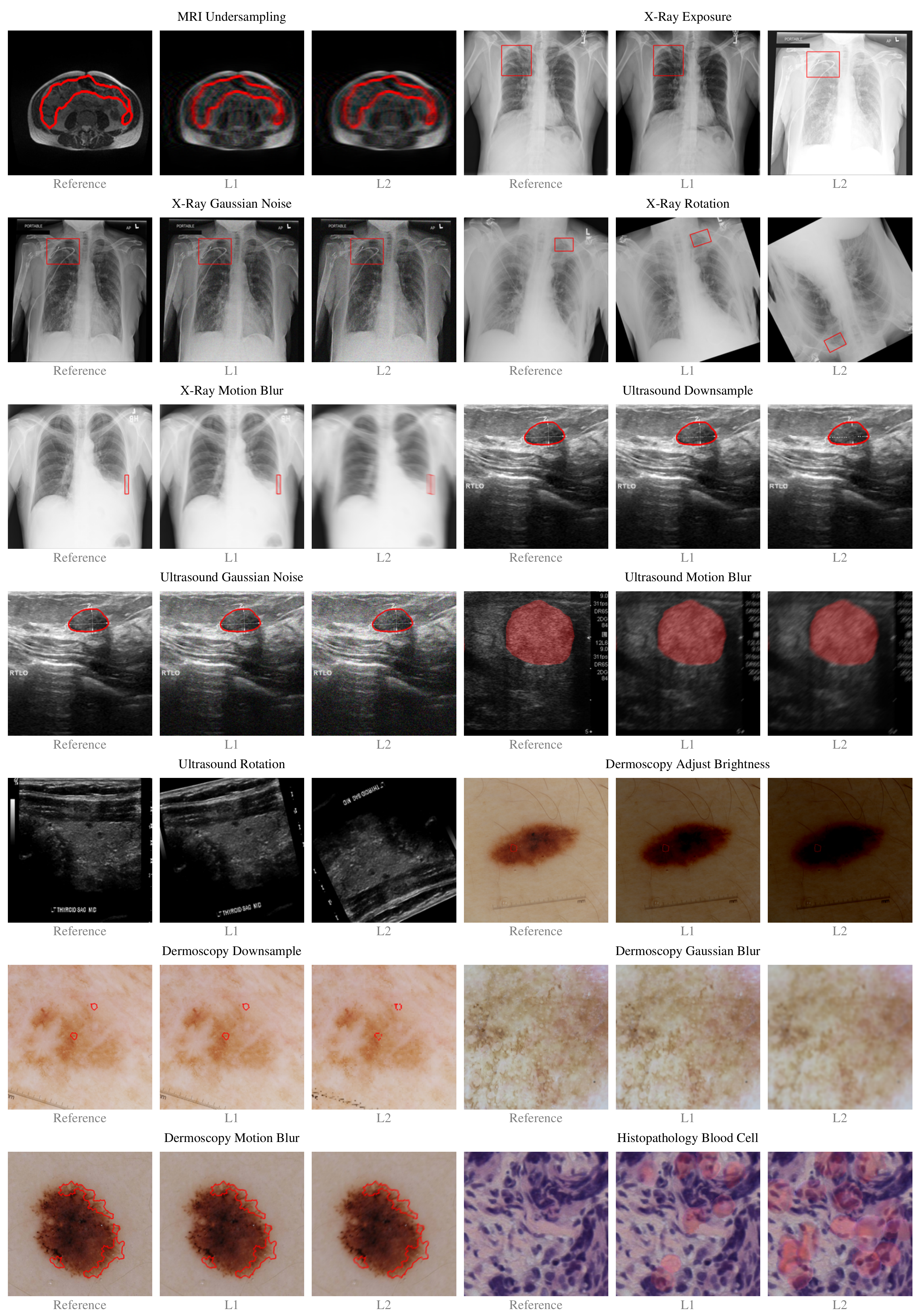}
    \caption{\textbf{Degradation examples (Part 2/3): X-ray, Ultrasound, and Dermoscopy modalities.}}
    \label{fig:degradation_examples_p2}
\end{figure}

\begin{figure}[!ht]
    \centering
    \includegraphics[width=\linewidth]{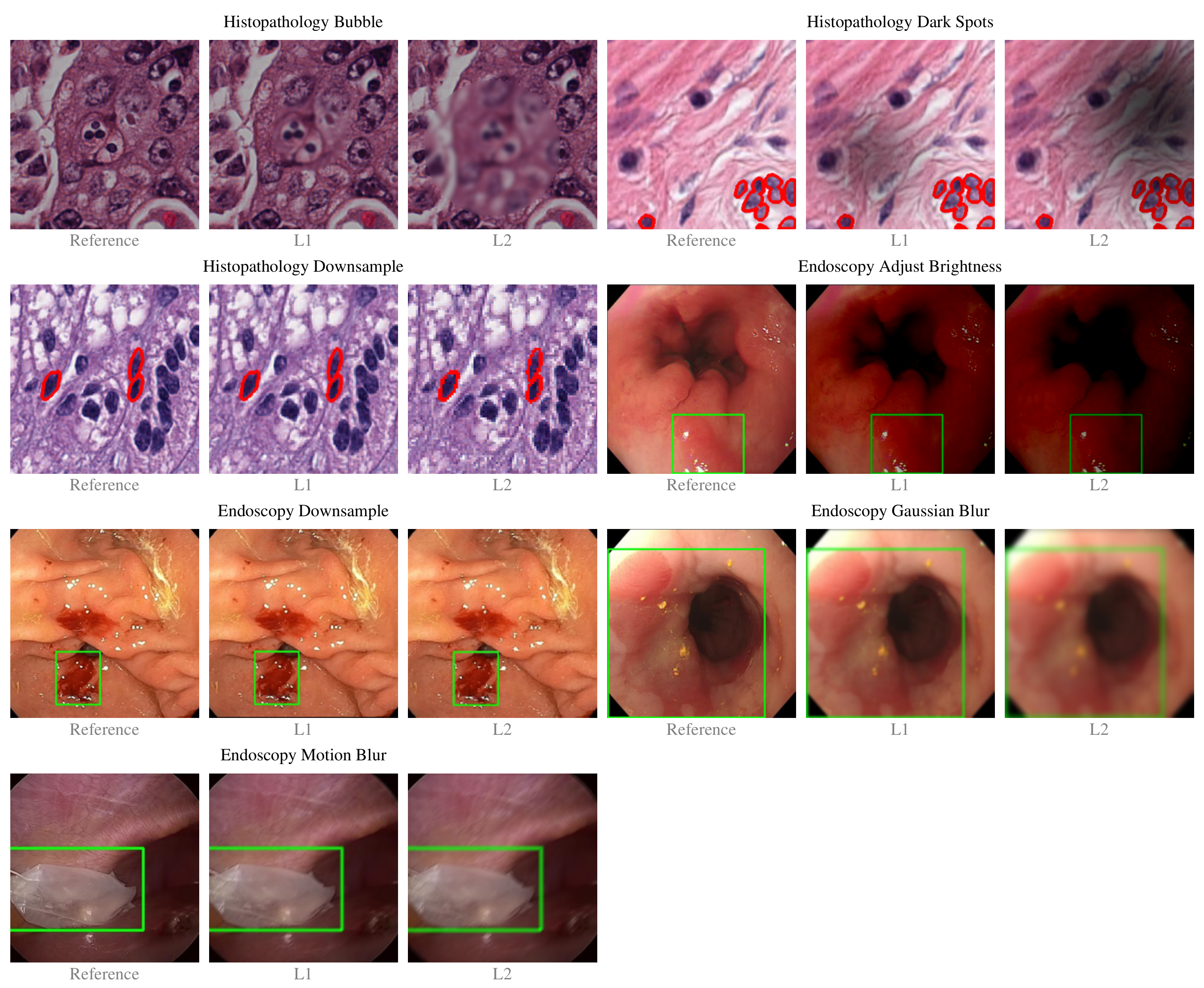}
    \caption{\textbf{Degradation examples (Part 3/3): Histopathology and Endoscopy modalities.}}
    \label{fig:degradation_examples_p3}
\end{figure}

Figures~\ref{fig:degradation_examples_p1}--\ref{fig:degradation_examples_p3} present representative visual examples of medical image quality degradations across different modalities and degradation types. Each row shows two degradation types, each applied at three severity degrees: Reference (L0), Mild (L1), and Severe (L2).

\end{document}